\documentclass[authoryear,preprint,review,12pt]{elsarticle}

\usepackage{amssymb}
\usepackage{amsmath}
\usepackage[linesnumbered,ruled,vlined]{algorithm2e}
\usepackage{array}
\usepackage{textcomp}
\usepackage{color}
\usepackage{subfigure}
\usepackage{stfloats}
\usepackage{url}
\usepackage{verbatim}
\usepackage{graphicx}
\usepackage{times}
\usepackage{latexsym}
\usepackage{algpseudocode}
\usepackage{booktabs}
\usepackage{graphicx}
\usepackage{threeparttable}
\usepackage{tabularx} 
\usepackage{hyperref}
\hypersetup{
colorlinks=true,
linkcolor=black,
citecolor=black,
urlcolor=black
}
\usepackage{amsthm}
\definecolor{newgreen}{RGB}{0,139,69}

\journal{Neurocomputing}

\begin{document}

\begin{frontmatter}



\title{Recursively Summarizing Enables Long-Term Dialogue Memory in Large Language Models} 

\author[a]{Qingyue Wang}
\ead{qingyue.wang@ust.hk}
\affiliation[a]{organization={the Department of Computer Science and Engineering, Hong Kong University of Science and Technology},
            city={Hong Kong},
            country={China}}
\author[c]{Yanhe Fu}
\ead{fuyanhe@iie.ac.cn}
\author[c]{Yanan Cao\corref{cor1}}
\cortext[cor1]{Corresponding author.}
\ead{caoyanan@iie.ac.cn}

\affiliation[c]{
organization = {the institute of information engineering, Chinese Academy of Sciences},
city={Beijing},
country={China}
}
\author[a]{
Shuai Wang}
\ead{shuaiw@cse.ust.hk}
\author[d]{Zhiliang Tian}
\ead{tianzhiliang@nudt.edu.cn}
\author[b]{Liang Ding}
\ead{liangding.liam@gmail.com}
\affiliation[d]{organization = {the College of Computer, National University of Defense Technology},
city={Changsha},
country={China}
}
\affiliation[b]{
organization={the University of Sydney},
city={Sydney},
country={Australia}
}

\begin{abstract}
Recently, large language models (LLMs), such as GPT-4, stand out remarkable conversational abilities, enabling them to engage in dynamic and contextually relevant dialogues across a wide range of topics. However, in a long-term conversation, these chatbots fail to recall appropriate information from the past, resulting in inconsistent responses. To address this, we propose to recursively generate summaries/ memory using large language models to enhance their long-term dialog ability. Specifically, our method first stimulates the LLM to memorize small dialogue contexts. After that, the LLM recursively produces new memory using previous old memory and subsequent
contexts. Finally, the chatbot is prompted to generate a response based on the latest memory. The experiments on widely used LLMs show that our method generates more consistent responses in long-term conversations, and it can be significantly enhanced with just two/ three dialog illustrations. Also, we find that our strategy could nicely complement both large context windows (e.g., 8K and 16K) and retrieval-enhanced LLMs, bringing further long-term dialogue performance. Notably, our method is a potential solution to enable the LLM to model the extremely long dialog context. We
release our code in \url{https://github.com/qingyue2014/Rsum}.
\end{abstract}


\end{frontmatter}



\section{Introduction}
Recently, large language models (LLMs), such as ChatGPT\footnote{\url{https://chat.openai.com/}} and GPT-4~\citep{GPT4OpenAI}, demonstrate promising performances in various natural language applications~\citep{Brown2020LanguageMA,Zeng2022GLM130BAO,zhong2023chat,Lu2023EAPrompt,Peng2023ChatGPT4MT,wu2023chatgpt}. One notable capability lies in their remarkable conversational prowess, comprehending input, and generating human-like responses. 

\begin{figure}[t]
  \centering
  \includegraphics[scale=0.4]{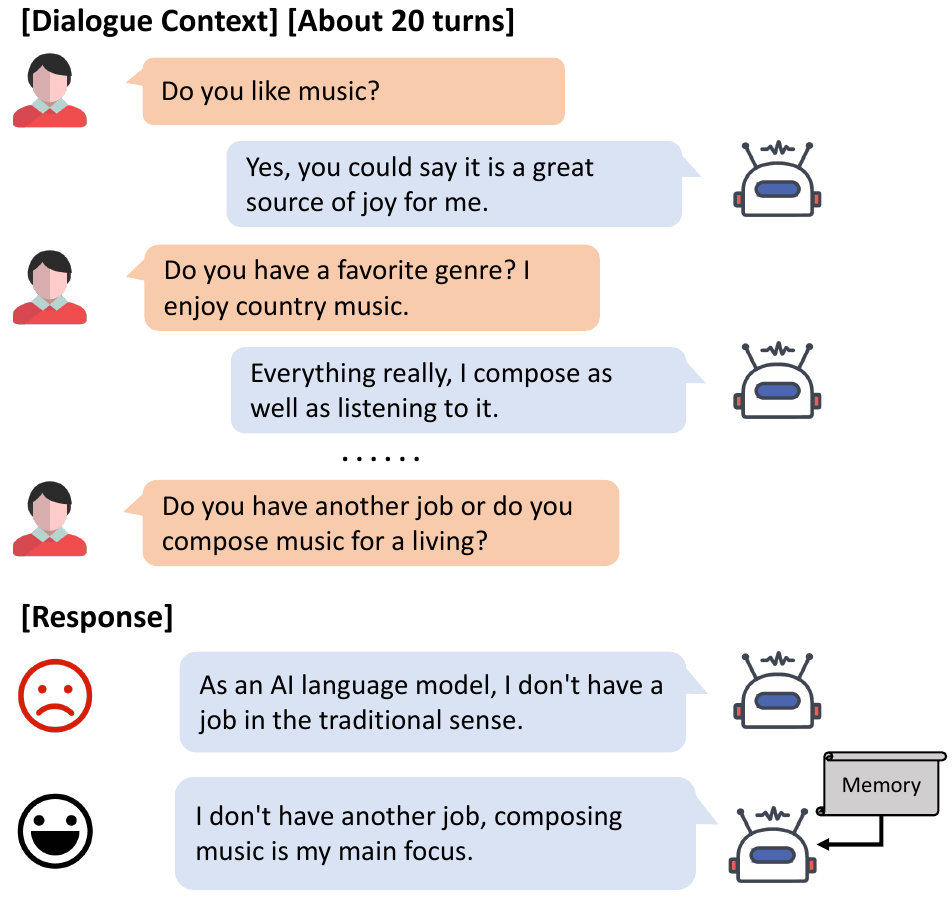}
  \caption{\textbf{A long-term conversation example} from the Multi-Session Chat Dataset~\citep{xu-etal-2022-beyond}. When the user refers back to previous subjects (i.e., composing music), even the ChatGPT (gpt-turbo-3.5-0301 version) generates an inconsistent response.}
  \label{fig:motivation}
\end{figure}

Large context windows allow many LLMs\footnote{Recent GPT-4 supports 128,000 tokens and LLama3\citep{meta2024llama3} supports 1,040,000 tokens.} to process entire dialog histories, yet they often struggle to effectively comprehend past interactions and integrate key information into responses~\citep{zhou-etal-2023-facilitating}. Applications such as personal AI companions, which need to recall past conversations for rapport building, and health assistants, which must consider a complete record of patient inquiries to provide diagnostic results, demonstrate the importance of maintaining consistency and coherence in long-term dialogues. Figure~\ref{fig:motivation} illustrates a dialog spanning over 20 turns, centered around a discussion of the speakers' personas (e.g., the bot composes music, and the user enjoys country music). However, even the powerful ChatGPT forgets past information and produces a poor response, showing the necessity to explicitly model long-term memory during conversations.

To address this, there are two mainstream methods to enhance the long-term dialog ability of LLMs. The first one is the retrieval-based method, which directly stores past conversational utterances in the storage and adapts an advanced retriever to identify the most relevant history~\citep{guu2020retrieval,Lewis2020RetrievalAugmentedGF}. However, it is difficult to obtain a well-performing (ideal) retriever, ensuring that the retrieved utterances capture the complete semantics about current conversations. The second way is to employ a memory module to summarize important conversation information to assist the LLM, which is also called memory-based approaches~\citep{mazare-etal-2018-training,longtimenosee, chen2024compress}. They usually apply a separately trained model or a powerful large language model to generate memory for past dialogues. 
Nevertheless, these methods lack the necessary iteration mechanism on generated memory, resulting in the reserved outdated information directly hurting the quality of responses.

In this paper, we propose a simple and effective plug-in method that enables LLM itself to generate summaries, which store the real-time information of speakers through continuous updating and reviewing past context to aid long-term interactions. In practice, a generative LLM is first prompted to produce a summary given a short dialog context. After that, we ask the LLM to continue updating and generate a new summary/ memory by combining the previous memory and subsequent dialogues. Finally, we encourage the LLM to respond using the latest memory as the primary reference to engage the ongoing dialogue. Given that the generated summaries are much shorter than the full dialogues, our proposed schema not only models long-term conversation memory but also serves as a potential solution to enable current LLMs to handle extremely long contexts (across multiple dialogue sessions) without expensively expanding the maximum length setting.

Experimentally, we implement our method using a variety of state-of-the-art open (Llama~\citep{llama2} and ChatGLM~\citep{glm2024chatglm}) and closed (OpenAI's GPT-3.5-Turbo) LLMs, and the performance on long-term dialog surpasses that of popular approaches both in automatic and human evaluations. Moreover, we verify the effectiveness of using explicit memory for long-term dialogs and using our generated memory is easier for LLMs to digest. These findings underscore the importance of developing advanced memory generation strategies. Our method can further enhance response quality by incorporating the in-context learning (ICL) technique, where multiple samples in the format of (dialogue, memory, and golden response) are presented to LLMs. This allows them to utilize the generated memory more flexibly. Additionally, we demonstrate the generalizability of our approach across different LLMs, with our method achieving approximately a +3\% improvement in BLEU score on text-davinci-003. Finally, we observe that our schema complements existing window-extended LLMs (e.g., GPT-3.5-Turbo-16k and LongLoRA-8k) and retrieval-enhanced LLMs (e.g., LLM-BM25 and LLM-DPR), producing more coherent and consistent responses in long-term conversations.

In summary, our \textbf{contributions} are as follows:
\begin{itemize}
    \item We propose a novel method by recursively summarizing past dialogues to enhance the LLM's memory, enabling the generation of highly consistent responses in long-term conversations.
    \item The solid experiments on the public datasets show the superiority of the proposed method, with multiple open-source and closed-source LLMs verifying its universality and robustness.
    \item The simplicity of our method makes it nicely complements existing works, including retrieval-based and long-context techniques, having the great potential to be an orthogonal plug-in for the LLM community.
\end{itemize}

\section{Related Work}
\subsection{Large Language Models}
Language language models (LLMs) have shown outstanding performance in a variety of user-facing language technologies, including conversation, summarization, and creative writing~\citep{GPT4OpenAI,Shuster2022BlenderBot3A, rubin2023long}. While these LLMs achieve notable success in many popular tasks, their ability to model long text remains a challenge~\citep{An2023LEvalIS}. To address the problem, some works are to adapt transformers to accommodate longer inputs, such as position interpolation~\citep{Chen2023ExtendingCW} and efficient self-attention~\citep{Beltagy2020LongformerTL, Chen2023LongLoRAEF}. However, these context window-extended LLMs not only require continual training on high-quality long texts but still struggle to use and retrieve the core information from the entire input~\citep{Liu2023LostIT}. Recently, in question-answer task, some researchers found that the performances of LLMs degrade significantly when people change the position of relevant information, indicating that current language models do not robustly make use of information in long input contexts~\citep{Liu2023LostIT,Li2023LooGLECL}. Many works suggest that the lack of explicit memory mechanisms in current LLMs hinders their performance on tasks requiring sustained context awareness and understanding~\citep{chen2024compress}. 
Nowadays, the performance of LLMs has not yet been explored deeply in long-range dialogue scenarios. This work focuses on developing the long-term modeling ability of the LLM, where we prompt it to self-memory, self-update, and self-use in conversations, aiding consistent response generation.
\subsection{Long-term Open-Domain Dialogue}
\begin{figure}[t]
 \centering
 \subfigure[Retriever method]{
 \includegraphics[scale=0.3] {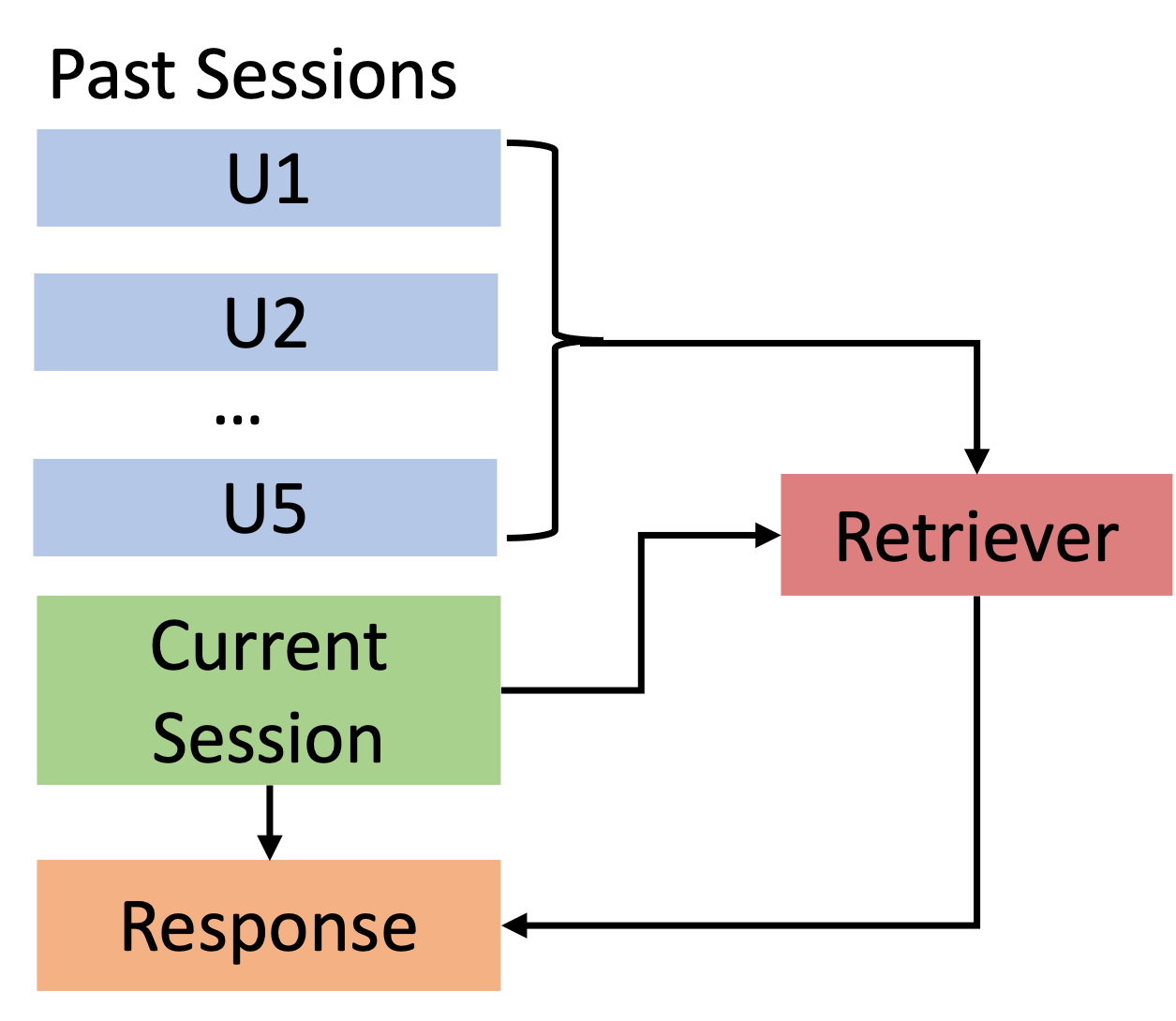}
 }\hspace{.05in}\subfigure[MemoChat]{
 \includegraphics[scale=0.3] {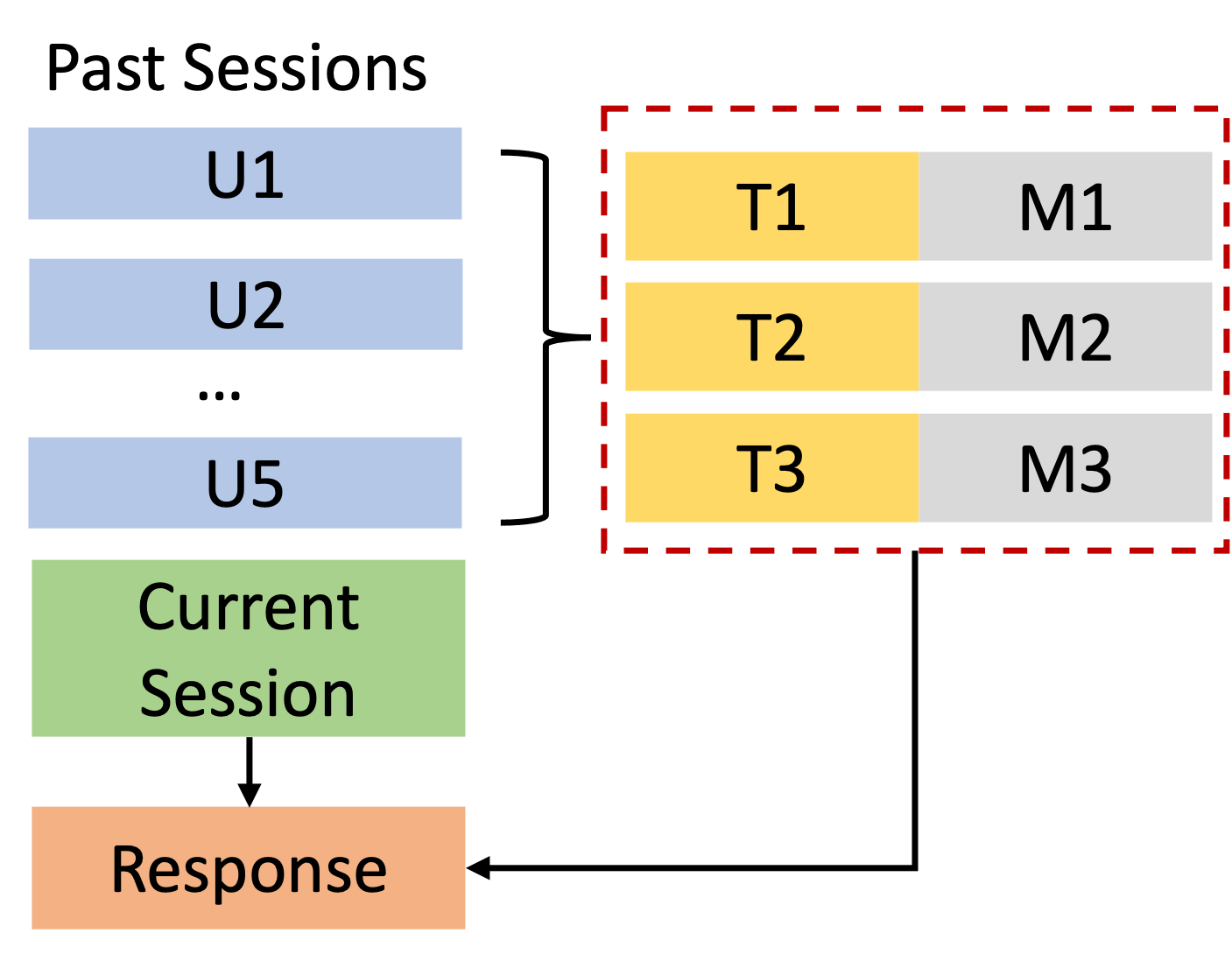}
 }\hspace{.05in}
 \subfigure[MemoryBank]{
 \includegraphics[scale=0.3] {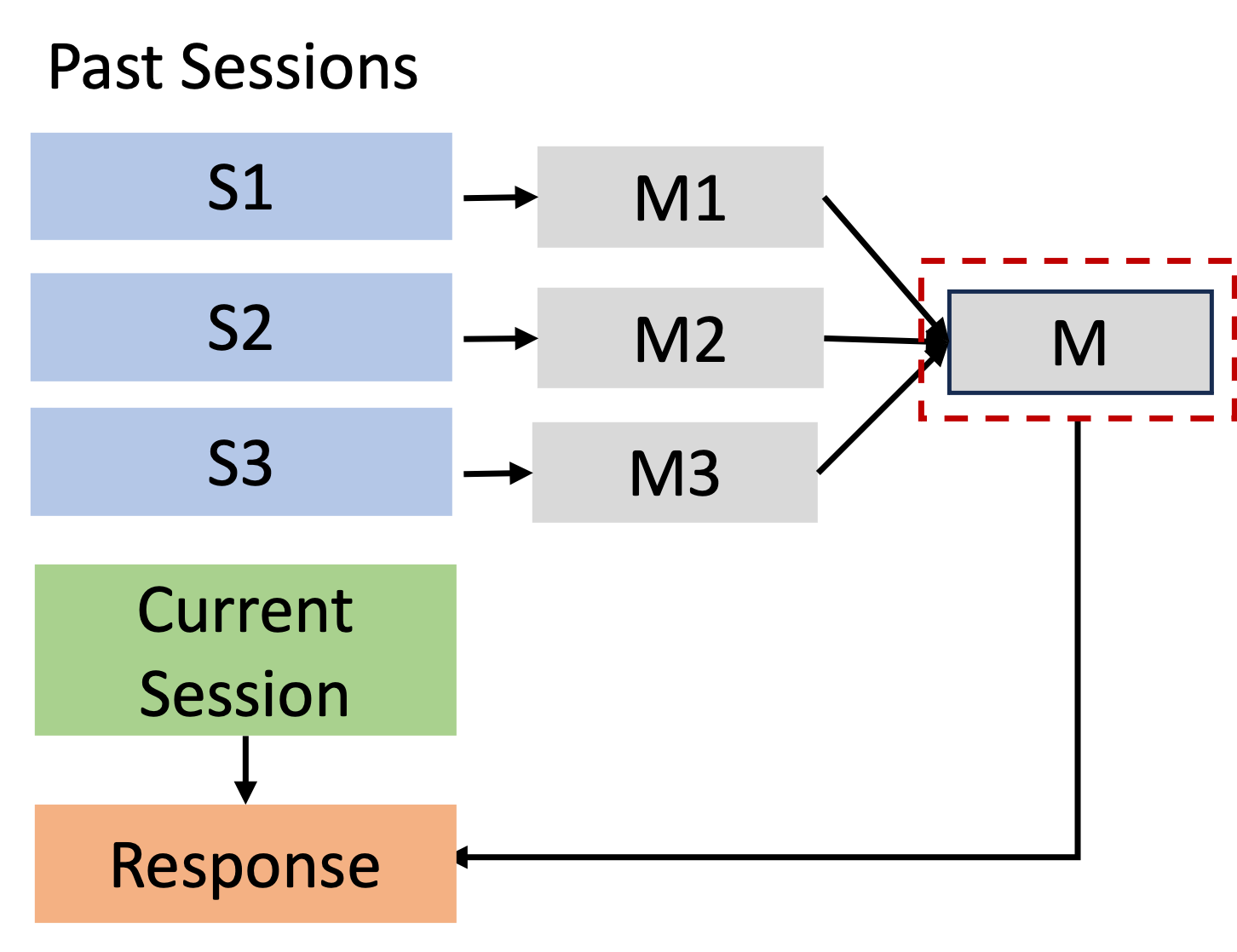}
 }\hspace{.05in}
 \subfigure[Ours]{
 \includegraphics[scale=0.3] {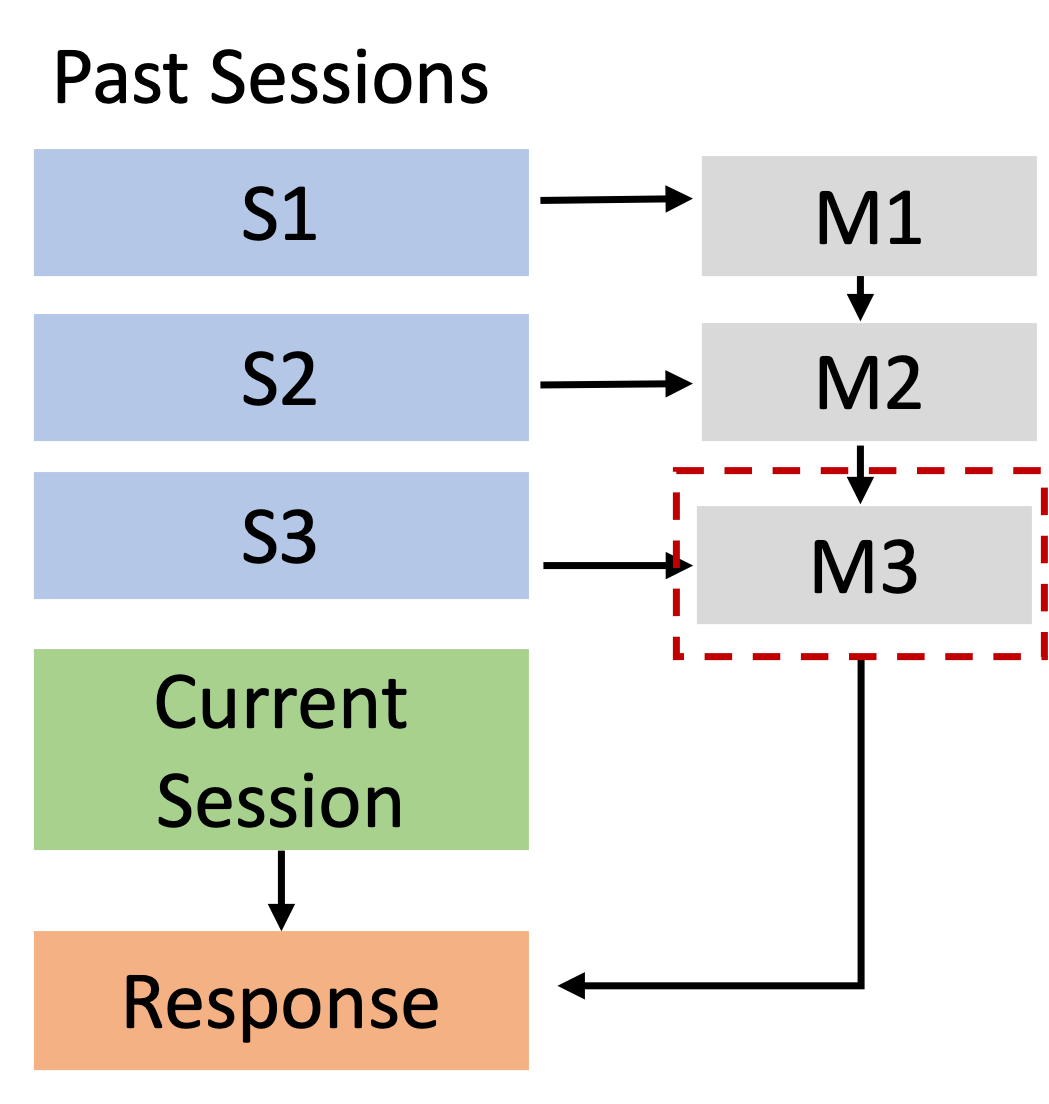}
 }
 \caption{\textbf{Comparison among baselines and ours.} The ``U'', ``S'', ``T'', and ``M'' are abbreviations for the Utterance, Session, dialog Topic, and Memory. The red dashed box refers to the memory used to generate the response.}
 \label{fig:com}
 \end{figure}

Open-domain dialogue systems~\citep{liu-etal-2016-evaluate, zhang-etal-2018-personalizing,Kann2022OpendomainDG}, also known as chatbots or conversational agents, have gained immense popularity and a lot of studies in recent years.
Among them, the long-term conversation setting is pretty a hard problem, because it needs the capability of understanding and memorizing key dialogue history information~\citep{wu-etal-2022-memformer,2022TongZhangHierarchical} about current query. The most popular and potential solution is to directly store the partial information for tracking the history of conversation~\citep{Lee2023PromptedLA}, usually in the form of dialogue utterances or summaries. The current conversation and relevant information are then inputted into the response generator. One intuitive idea is to apply a retriever to find the most relevant utterances according to the current dialog, which is called as the retrieval-based method. Another popular method is a memory-based method, which tries to generate and manage the summary to obtain key information from history. For example, MemoChat~\citep{Lu2023MemoChatTL} allows chatbots to reorganize the past dialogue histories according to different topics of speakers and prompt the LLM to retrieve from the structured memory during generation. Going further, MemoryBank~\citep{Zhong2023MemoryBankEL} proposes a new memory mechanism by generating summaries for each dialog session first and then compressing them into a global one. However, their memory is completely fixed once stored, failing to guarantee its consistency with ongoing dialog. The important comparison between these existing methods and ours is shown in Figure~\ref{fig:com}. As seen, our approach and these methods mainly diverge in the way of memory generation, where we continuously integrate historical information and old memory to obtain real-time memory, enabling the gain of accurate memory and modeling of long-distance dependencies. 

\section{Approach Overview}
Following previous works~\citep{xu-etal-2022-beyond, bae-etal-2022-keep}, we denote that a long-context dialogue consists of multiple sessions with a specific user, which is also called \textit{Multi-Session Dialogue}. The goal of the task is to generate context-relevant and highly consistent responses to the user based on past sessions and current context. Formally, each dialogue can be written as $D=\{S,C_t,r_t\}$. Here, $S=\{S_1, S_2, ...,S_N\}$ represents $N$ past sessions and each of the sessions consists of multiple utterances between two speakers. $r_t$ is the ground truth response to $C_t$ with background sessions $S$. $C_t=\{u_1,r_1,...,u_t\}$ denotes the dialogue context of the current session at $t$ step, where $u$ and $r$ represent the utterances from the user and the chatbot, respectively.  

In this paper, we propose a new memory mechanism to aid a large language model for multi-session dialog tasks. The memory contains multiple natural language sentences, storing the key information of speakers extracted from previous sessions. Our goal is to obtain a reliable memory given past sessions and predict a consistent and meaningful response using the current dialogue context and the memory. Specifically, we decompose the goal into two stages with the following probability distribution:
\begin{equation}
    P(r_t|C_t,S) = P(r_t|C_t,M_N)P(M_N|S),
\end{equation}
where $M_i$ represents the available memory when the $i$-th session is finished. And $P(M_N|S)=\prod_{i=1}^{N}P(M_i|S_i,M_{i-1})$ is a sequential or Markov process where each memory $M_i$ of session $i$ depends only on the current session and the previous memory $M_{i-1}$.

\begin{figure*}[t]
  \centering
  \includegraphics[scale=0.65]{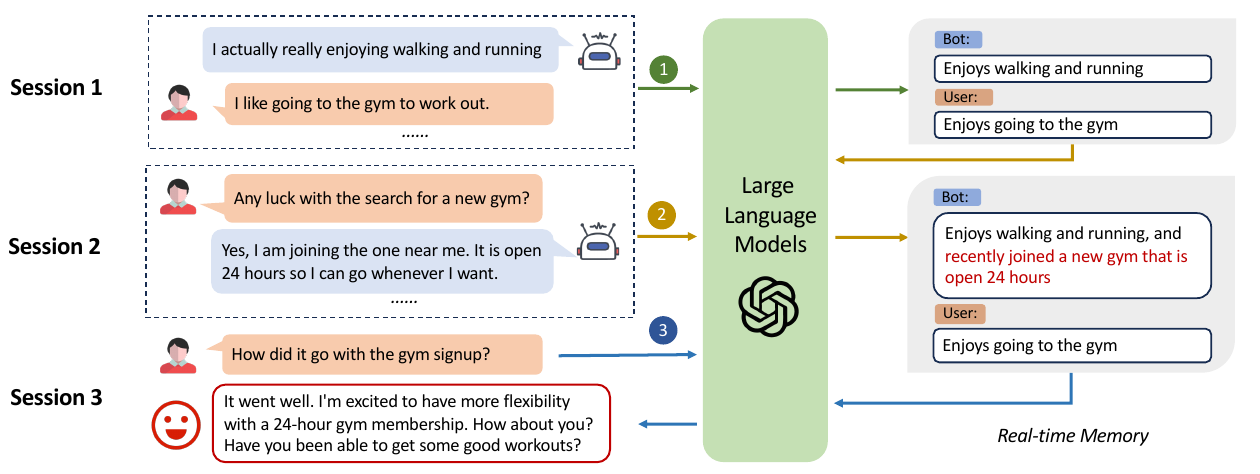}
  \caption{\textbf{The schematic overview of our method}. The model uses the first session to generate initial memory (green arrows), then updates the memory when the second session ends (yellow arrows), and generates a response using the latest memory at the third session (blue arrows).}
  \label{fig:model}
\end{figure*}
\section{Approach}
To achieve long-term dialog, we prompt an arbitrary large language model to finish two tasks, i.e., \textbf{memory iteration} and \textbf{memory-based response generation}. The former is responsible for recursively summarizing the key information along with long-term dialogue, and the latter is to incorporate the latest memory and current dialog to generate an appropriate and consistent response. The workflow of our proposed method is shown in Figure~\ref{fig:model}.

\subsection{Memory Iteration}
\label{sec:memory_iter}
The goal of memory iteration is to obtain a coherent and up-to-date summary for the chatbot.
Early works~\citep{bae-etal-2022-keep,choi2023effortless} update memory by carrying multiple ``hard operations'' on summaries, such as \textit{replace}, \textit{append}, and \textit{delete}, which rely on high-quality dialogue with operation labels. However, this laborious design disrupts the semantic coherence of the summary and is not suitable for management over a long period. Differently, we guide the LLMs to recursively self-generate memory (summaries) using dialogue context and previous memory. By utilizing old summaries, the model can fully digest the current dialog context and thus gain a high-quality memory. Formally, the updated memory is computed by:
\begin{equation}
    M_i=\boldsymbol{LLM}(S_{i},M_{i-1},\mathrm{P}_{m}),
\end{equation}
where $M_i=\{m_1, m_2, ..., m_J\}$ denotes multiple sentences, containing summarized key information from the session $S_i$, and $\mathrm{P_m}$ is the prompt of LLM for generating new memory. The memory iteration will be repeated $N$ times until all previous sessions end, where we can obtain the latest memory $M_N$. Take the dialog in 
 Figure~\ref{fig:model} as an example, two memory iterations happen at the end of the first and second sessions. In the second iteration, the LLM incorporates the new personality (i.e., the bot recently joined a new gym) from Session 2 into the old memory (i.e., the bot enjoys walking and running).
\paragraph{Prompt Construction} To enable LLM to efficiently carry the memory iteration task, we design a specific prompt for it, which is shown in Table~\ref{tab:memory_prompt}. It mainly consists of three parts: (1) \textbf{Task definition} is responsible for defining the role of the current LLM, as well as the memory iterator's (LLM) input and output. (2) \textbf{Task description} gives detailed steps to finish the above task. To make sure the memory update is timely, we remind the LLM to create a new representation of speakers by considering old summaries and current sessions.(3) \textbf{Task input} contains two placeholders, where we take previous memory and a whole session as the inputs. Through experimental verification, we found that using step-by-step instructions helps the LLM better understand and execute the memory iteration\footnote{\url{https://platform.openai.com/docs/guides/prompt-engineering}}.

\begin{table}[t]
\centering
\caption{\textbf{The prompt design of memory iteration}, including \textcolor{blue}{task definition}, \textcolor{newgreen}{task description} and \textcolor{red}{task inputs}}
\label{tab:memory_prompt}
\scalebox{0.68}{
\begin{tabular}{ll}
\toprule
\textbf{Prompt} & \begin{tabular}[c]{@{}l@{}}\textcolor{blue}{You are an advanced AI language model with the ability to store and update a memory to keep track of }\\ \textcolor{blue}{key personality information for both the user and the bot. You will receive a previous memory and }\\ \textcolor{blue}{dialogue context. Your goal is to update the memory by incorporating the new personality information.}\\ \textcolor{newgreen}{To successfully update the memory, follow these steps:}\\ \textcolor{newgreen}{1.Carefully analyze the existing memory and extract the key personality of the user and bot from it.}\\ \textcolor{newgreen}{2. Consider the dialogue context provided to identify any new or changed personality that needs to }\\ \textcolor{newgreen}{be incorporated into the memory.}\\ \textcolor{newgreen}{3. Combine the old and new personality information to create an updated representation of the user and } \\\textcolor{newgreen}{bot's traits.}\\ \textcolor{newgreen}{4. Structure the updated memory in a clear and concise manner, ensuring it does not exceed 20 sentences.}\\  
\textcolor{newgreen}{Remember, the memory should serve as a reference point to }\\\textcolor{newgreen}{maintain continuity in the dialogue and help you respond accurately to the user based on their personality.}
\\
\textcolor{red}{{[}Previous Memory{]} {[}Session Context{]}}\end{tabular} \\
\midrule
\textbf{Output} & {[}Updated Memory{]}  \\ \bottomrule         
\end{tabular}}
\end{table}

\subsection{Memory-based Response Generation}
The final goal is to produce consistent and natural responses given dialogue memory and the context of the current session. Formally, the response of the current session can be obtained by stimulating the LLM as a response generator: 
\begin{equation}
    r_t=\boldsymbol{LLM}
(C_t,M_N,\mathrm{P_r}),
\end{equation}
where $P_r$ is a prompt of the generator. Especially, taking the generated memory $M_N$ and current session $C_t$, we ask the LLM again to generate a response. In Figure~\ref{fig:model}, the LLM considers the newest memory from Session 2 and provides a high consistent response to the user, i.e., `` more flexibility with a 24-hour gym membership''.
\begin{table}[t]
\centering
\caption{\textbf{The prompt of memory-based response generation}, including \textcolor{blue}{task definition}, \textcolor{newgreen}{task description} and \textcolor{red}{task inputs}}
\label{tab:generator}
\scalebox{0.68}{
\begin{tabular}{ll}
\toprule
\textbf{Prompt} & \begin{tabular}[c]{@{}l@{}}\textcolor{blue}{You will be provided with a memory containing personality information for both yourself and the user. } \\ \textcolor{blue}{Your goal is to respond accurately to the user based on the personality traits and dialogue context.} \\ \textcolor{newgreen}{Follow these steps to successfully complete the task:}\\ \textcolor{newgreen}{1. Analyze the provided memory to extract the key personality }  \textcolor{newgreen}{traits for both yourself and the user.}\\ \textcolor{newgreen}{2. Review the dialogue history to understand the context and flow} \textcolor{newgreen}{of the conversation.} \\ \textcolor{newgreen}{3. Utilize the extracted personality traits and dialogue context to} \textcolor{newgreen}{formulate an appropriate response.} \\ \textcolor{newgreen}{4. If no specific personality trait is applicable, respond naturally} 
\textcolor{newgreen}{as a human would.}\\ 
\textcolor{newgreen} {5. Pay attention to the relevance and importance of the personality information, focusing on capturing} \\
\textcolor{newgreen}{the most significant aspects while maintaining the overall coherence of the memory.}
\\
\textcolor{red}{{[}Previous Memory{]}  {[}Current Context{]}}\end{tabular} \\ \midrule
\textbf{Output} & {[}Response{]}  \\ \bottomrule
\end{tabular}}
\end{table}

\paragraph*{Prompt Construction} The the prompt for memory-based response generation is shown in Table~\ref{tab:generator}. The prompt is similar to that of memory iteration, including task definition, description, and inputs, where we remind the LLM to utilize the extracted information and maintain the consistency of memory when responding. Also, the step-by-step instructions method is also effective for memory-based response generation.

\begin{algorithm}[t]
\caption{Response generation using recursive memory.}
\label{alg:inference}
\KwIn{A long-term dialog $D=\{S,C_t\}$ consisting multiple sessions with a user; A generative pre-trained model LLM; Pre-defined prompt $P_m$ and $P_r$}
\KwOut{A response to user.}
$M_0 \leftarrow \text{none}$\;\tcp{Set initial memory as empty}
\For{$i \leftarrow 1$ \KwTo $N$}{
    $M_i = LLM(S_i, M_{i-1}, P_m)$\;\tcp{Update memory when a session ends}
}
 $r_t=LLM(C_t, M_N, P_r)$\;\tcp{Response using latest memory}
\KwRet{$r_t$}
\end{algorithm}
\subsection{Algorithm}
The process of response generation using recursive memory is illustrated in Algorithm~\ref{alg:inference}. In the beginning, the initial memory is set as an empty, i.e., ``none'' string. After that, we recursively update the memory using each session context (line 3). Finally, the LLM generates a response with the help of the latest memory (line 5). The generative pre-trained models used for memory iteration and response generation can be different. For instance, the developers can train a private memory iterator through customized models or data, to enhance the target open/ closed LLMs for long-term or long-context tasks.

\section{Experimental Settings}
\subsection{Datasets}
We validate the effectiveness of the proposed method on two widely-used long-term dialogue datasets: Multi-Session Chat (MSC) dataset~\citep{xu-etal-2022-beyond} and Carecall dataset ~\citep{bae-etal-2022-keep}. \paragraph{MSC dataset} is the largest human-human long conversations dataset so far. The early sessions are a short conversation where two
speakers get to know each other for the first time and then they either continue to talk about the previous subject or spark up conversation on a new topic. \paragraph{Carecall} is a Korean open-domain multi-session dataset, which is used for monitoring patient health. For a fair comparison, we use the public machine-translated English version\footnote{\url{https://github.com/naver-ai/carecall-memory}} in the experiments. 

The CareCall's setting is similar procedure presented in the MSC dataset. 
The main difference is that the Carecall additionally contains more persona updates that are likely to change in a short period, such as the user's health and diet, while the persona information in the MSC dataset remains fixed once it is stored. Both of the two datasets have five sessions and each one consists of multiple utterances between two speakers (the user and chatbot), where the conversationalists reengage after several hours or days and continue chatting. As early sessions only have a very short history of conversations, we mainly evaluate the proposed method in session 4 and 5 for the ability of long-term modeling. The statistics of two datasets are given in ~\ref{ap:dataset}. 

\subsection{Evaluation Metrics}
we conduct diverse evaluations during experiments, including automatic metrics, human evaluation, and LLM judgments, focusing on the quality of generated memory and response. 
\paragraph*{Automatic Metrics}
We employ BLEU-1/2~\citep{BLEUAMethod}, F1 with reference to the human annotations. Besides, we compute BertScore~\citep{li-etal-2016-diversity} to measure the semantics similarity between the references and the generated responses. 
\paragraph*{Human Evaluation} Many works point out that automatic evaluation metrics are insufficient for capturing the nuances of conversations ~\citep{Deriu2019SurveyOE}. Following the previous works~\citep{bae-etal-2022-keep}, we ask three crowd-sourcing workers to assign a score from 0 to 2 (0:bad, 1:OK, 2:good) to the generated responses based on the aspects of engagingness, coherence, and consistency. These criteria are discussed as follows: (1) \textit{Engagingness}: It evaluates whether the chatbot captures the user's interest and makes them want to continue the conversation. A high score on engagingness means the responses are interesting and contextually appropriate, encouraging users to keep chatting.
(2) \textit{Coherence}: It measures whether the response maintains a logical and clear flow based on the conversation's context. A coherent response ensures the conversation makes sense and stays relevant, enhancing user engagement. (3) \textit{Consistency}:
It assesses whether the response aligns with the information provided in previous interactions. Consistent responses build trust and reliability by demonstrating that the chatbot remembers and integrates past exchanges accurately.

\paragraph*{LLM Evaluation} Recently, LLM-as-a-Judge strategy~\citep{Pan2023DoTR} has been widely used in evaluating generation tasks. Some works reveal minimal deviation of GPT-4’s evaluation from humans (\textgreater0.85 agreements) in dialog quality~\citep{Zhang2023ACA}. Inspired by this, we employ the GPT-4 as an advanced evaluator, using two common methods to assess the quality of generated responses. (1) \textit{Single model evaluation}~\citep{Lu2023MemoChatTL}: we prompt GPT-4 to rate the responses individually from the three aspects, i.e., engagingness, coherence and consistency with an integer scale from 1 (very bad) to 100 (very good). (2) \textit{Pairwise model evaluation}~\citep{Dubois2023AlpacaFarmAS}: we ask the GPT-4 to directly compare two anonymous generations and determine which response is better. While single model evaluation provides detailed insights into specific aspects of each response, pairwise comparison is essential for understanding relative performance, particularly when distinguishing subtle differences between outputs.
\subsection{Baselines}
We mainly employ the following methods for long-text dialogues in LLMs: context-only approaches (without using any memory), retrieval-based approaches (with different retrievers), and memory-based approaches (with different memory mechanisms)
\paragraph*{Context-only Approach} It is the most naive approach to directly employ the LLM as a chatbot, where it concatenates past sessions and current dialogue context as the input. We use ``Llama2-7B''~\citep{llama2}, ``ChatGLM2-6B''\footnote{\url{https://github.com/THUDM/ChatGLM2-6B}}, and OpenAI ChatGPT ``gpt-3.5-turbo-0301'' as the backbone LLMs for the context-only approach\footnote{For convenience's sake, the following ``ChatGPT'' refer to the gpt-3.5-turbo-0301 version.}. 
\paragraph*{Retrieval-based Approach}
    Many previous works~\citep{xu-etal-2022-beyond} employ retrievers to filter key information and then include top-$k$ documents into inputs to assist long-context dialogs. For the long-term dialog, the top-$k$ documents refer to the relevant utterances from history. Here, we choose two widely used retrieval algorithms, i.e., BM25~\citep{robertson2009probabilistic} and pre-trained dense passage retrieval (DPR)~\citep{karpukhin-etal-2020-dense}, to look up the relevant utterances from past sessions. For convenience, we name the above retrieval-based baselines as \textbf{ChatGPT-BM25} and \textbf{ChatGPT-DPR}, respectively.
 \paragraph*{Memory-based Approach} Recent works employ a summarizer to abstract important information from the past to aid long-term conversation. Simply, we just choose two representative methods from various memory-based techniques, MemoryBank~\citep{Zhong2023MemoryBankEL} and MemoChat~\citep{Lu2023MemoChatTL}. \textit{MemoryBank} proposes a human-like long-term memory mechanism, which creates ordered summaries of past dialogs with timestamps, and then reorganizes them to obtain the global memory. The memory will be forgotten and updated by Ebbinghaus’s forgetting curve. Here, we plug MemoryBank with ChatGPT as a strong baseline, named \textbf{ChatGPT-MemoryBank}. Differently, \textit{MemoChat} maintains the structured conversational memory to aid long-term dialogue, i.e., generating the summaries for each dialogue topic. We plug the MemoChat into ChatGPT, named \textbf{ChatGPT-MemoChat}, for a fair comparison with others.

Note that our approach focuses on the zero-shot setting for LLMs to engage in a long-term dialog, making the comparisons with other fine-tuned models unfair.

\subsection{Implementation}
We implement our method by letting the LLM response using recursively generated memory in a long-term dialogue, thus it is called as ``\textbf{LLM-Rsum}''. 

\paragraph*{Backbone LLMs}We employ OpenAI ChatGPT ``\textit{gpt-3.5-turbo-0301}'', ``\textit{Llama2-7B}'' and ``\textit{ChatGLM2-6B}'' in the main experiments, ``\textit{text-davinci-003}'' and ``\textit{Llama2-7B}''~\citep{llama2} in the analysis to show the universality, ``\textit{longlora-8k}''~\citep{Chen2023LongLoRAEF} and ChatGPT-16k ``\textit{gpt-3.5-turbo-16k}'' as the backbones of complementary discussion. Unless otherwise specified, we employ the same LLM to finish the memory iteration and memory-based response generation. During generation, we set the temperatures of all LLMs as 0 for fair comparisons. The max length of input tokens for Carecall and MSC datasets is no more than 4k, thus all backbone LLMs in experiments can process the entire dialog context. 

\paragraph*{Retrievers} Considering the scale of past utterances is not large enough to use the FAISS~\citep{faiss2019}, we choose the top-k most relevant utterances that will be combined with ongoing dialogues, prompting the LLM to respond. Following previous works~\citep{xu-etal-2022-beyond} for long-term dialogs, the k is set into 3 and 5.
\paragraph*{Memory-based Approaches}The implementation and prompt design of the MemoryBank and Memochat methods are based on the code publicly released in their original papers. For details, please refer to the original papers.
\paragraph*{LLM evaluations} We use the GPT-4 model (version \textit{``gpt-4-0314''}) as the evaluator, setting the temperature to 0 when making judgments. The prompts for evaluating the single model and pairwise models are referred to \citep{Lu2023MemoChatTL} and \citep{Dubois2023AlpacaFarmAS}, respectively. All prompts used in the experiments can be seen in~\ref{ap:prompt}.
\section{Experimental Results}
\subsection{Main Results} 
\begin{table}[t]
\centering
\caption{\textbf{Comparison of automatic and human evaluations among different methods} on MSC and Carecall datasets, reporting the quality of generated response. The ``BScore'', ``Enga.'', ``Cohe'' and ``Cons'' are the abbreviations of BertScore, Engagingness, Coherence, and Consistency. The best value is \textbf{bolded}.}
\label{tab:msc}
\resizebox{\linewidth}{!}{
\begin{tabular}{lllllllllllll}
\toprule
                                  & \multicolumn{6}{c}{\textbf{MSC Dataset}}                                                               & \multicolumn{6}{c}{\textbf{Carecall Datset}}                                                           \\
\cmidrule(lr){2-7} \cmidrule(lr){8-13}

Method                            & F1             & BLEU-1/2             & BScore         & Enga.         & Cohe.         & Cons.         & F1             & BLEU-1/2             & BScore         & Enga.         & Cohe.         & Cons.         \\
\midrule
\textit{Context-only LLM}              &                &                      &                &               &               &               &                &                      &                &               &               &               \\
Llama2-7B                         & 16.43          & 20.96/12.09          & 84.04          & 1.32          & 1.20          & 1.13          & 13.71          & 20.89/12.28          & 84.49          & 0.75          & 0.75          & 1.00          \\
ChatGLM2-6B                       & 15.38          & 21.69/12.51          & 84.48          & 1.10          & 1.15          & 1.07          & 13.09          & 20.59/12.03          & 84.91          & 0.66          & 0.63          & 0.86          \\
ChatGPT                           & 19.41          & 21.23/12.24          & 86.13          & 1.83          & 1.37          & 1.32          & 13.69          & 21.15/12.20          & 85.53          & 1.50          & 1.52          & 1.43          \\
\midrule
\textit{Retrieval-based Approach} &                &                      &                &               &               &               &                &                      &                &               &               &               \\
ChatGPT-BM25 (k=3)                      & 19.56          & 21.60/12.46          & 85.82          & 1.72          & 1.48          & 1.32          & 12.64          & 21.57/12.44          & 85.24          & 1.40           & 1.31          & 1.31          \\
ChatGPT-DPR  (k=3)                     & 20.23          & 21.75/12.55          & 86.04          & 1.76          & 1.51          & 1.34          & 12.21          & 21.39/12.35          & 85.25          & 1.55          & 1.35          & 1.45          \\
\midrule
\textit{Memory-based Approach}    &                &                      &                &               &               &               &                &                      &                &               &               &               \\
ChatGPT-MemoChat                  & 18.93          & 21.82/\textbf{12.59}          & 85.99          & 1.70          & 1.55          & 1.35          & 11.19          & 21.07/12.18          & 85.22          & 1.45          & 1.20          & 1.30          \\
ChatGPT-MemoryBank                & 20.28          & 21.82/12.58          & 86.12          & 1.78          & 1.57          & 1.40          & 13.15          & 21.29/12.39          & 85.34          & 1.57          & 1.52          & 1.68          \\
ChatGPT-Rsum (Ours)                & \textbf{20.48} & \textbf{21.83/12.59} & \textbf{86.89} & \textbf{1.85} & \textbf{1.60} & \textbf{1.45} & \textbf{14.02} & \textbf{21.64/12.48} & \textbf{86.05} & \textbf{1.62} & \textbf{1.60} & \textbf{1.70}\\
\bottomrule
\end{tabular}
}

\end{table}
\paragraph*{Automatic Metrics Results} In Table~\ref{tab:msc}, we compare different methods over session 5 in the MSC and Carecall datasets using popular LLMs. Firstly, among the vanilla models (``Llama2-7B'', ``ChatGLM2-6B'', and ``ChatGPT''), ChatGPT consistently performs well across two datasets, with competitive scores in BScore, F1 and BLEU-1/2. 
These results illustrate that ChatGPT is robust enough to handle a long-term dialog, thus, we leave ChatGPT as the backbone model for our method.
Secondly, as expected, the proposed method (``ChatGPT-Rsum'') achieves the best performance on both datasets, showing the benefits of using automatically recursive memory. Specifically, our method achieves about +0.2\% on the F1 score, which is acceptable compared to that in previous works~\citep{xu-etal-2022-beyond}. The MSC datasets are harder than other open-domain datasets due to the 3x context, so the slight improvements are normal. 
Thirdly, retrieval-based methods may not be always helpful in enhancing the quality of generation. From the results, the performances of ChatGPT-BM25 and ChatGPT-DPR are much worse than vanilla ChatGPT in the Carecall dataset, which is completely contrary to that in MSC. The reason is that the chatbot needs to actively guide the dialog topics in the Carecall, thus it is hard to retrieve appropriate and relevant context from the user's query. Therefore, the performance of generated responses will be damaged due to unrelated information.

\paragraph*{Human Evaluation Results} We also present the results of human evaluations on different methods in Table~\ref{tab:msc}.
From the results, we find that: 1) Most memory-augmented methods gain higher scores on consistency and coherence than vanilla ChatGPT, proving that maintaining a memory is more effective than using the whole history directly when engaging in a long-term conversation for LLMs; 2) Our method can generate more engaging response than other memory-based baselines (ChatGPT-MemoryBank and ChatGPT-MemoChat). The reason is that continually updating memory actively establishes global dependencies inside past histories, which helps LLMs to better understand the dialog and generate high-quality responses. 

\subsection{LLM Evaluation} 
\begin{figure}[t]
  \centering
\includegraphics[scale=0.5]{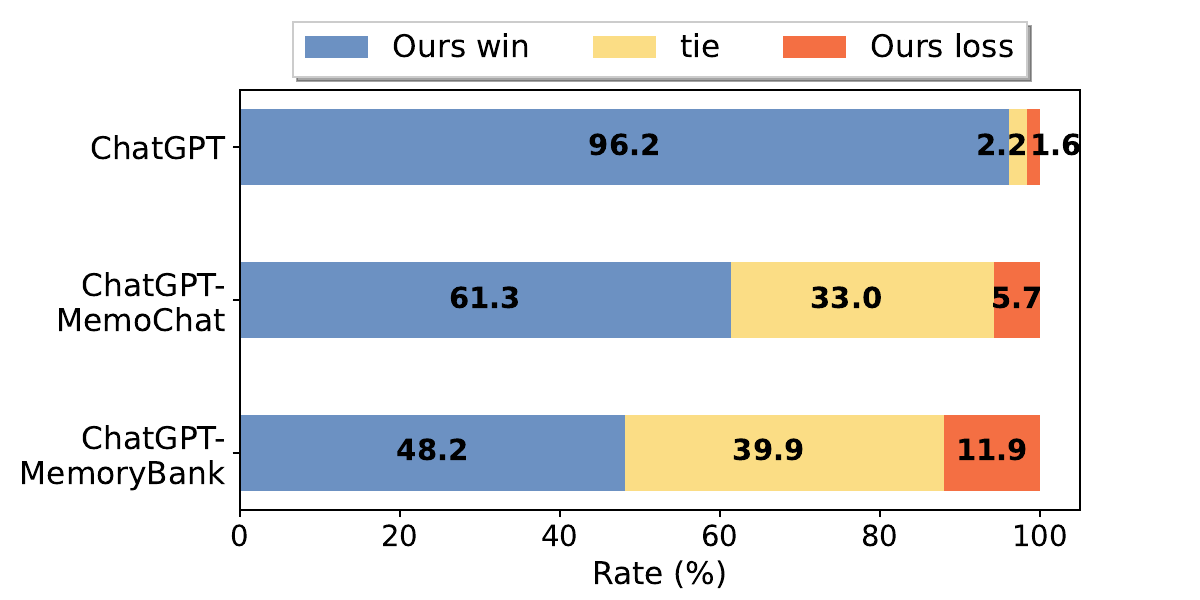}
  \caption{\textbf{Comparative win rate of our method and competitive baselines,} including ChatGPT, ChatGPT-MemoChat, and ChatGPT-MemoryBank.}
  \label{fig:win}
\end{figure}

\paragraph*{Single Model Evaluation} Table~\ref{tab:gpt4} reports the GPT-4's evaluation metrics results on session 5 among various methods in the MSC dataset. Also, the results prove the high agreements between humans (in Table~\ref{tab:msc}) and GPT-4 judgment on the overall quality of generated responses, i.e., ChatGPT-Rsum \textgreater ChatGPT-MemoryBank \textgreater  ChatGPT-MemoChat \textgreater  ChatGPT. Given this, we mainly present the LLM evaluations in the following experiments to reduce labor costs. Lastly, it is worth noting that compared to human evaluation results, the GPT-4 tends to give higher scores on both sentence coherence and consistency. We suppose that the values of humans and LLMs might not be fully aligned at a fine-grained level, which could be a new direction for developing LLM evaluation.
\begin{table}[t]
\centering
\caption{\textbf{Comparison of evaluation metrics results by GPT-4} across different methods on session 5 in MSC dataset.}
\label{tab:gpt4}
\resizebox{0.75\linewidth}{!}{
\begin{tabular}{lllll}
\toprule
Method & Engagingness & Coherence & Consistency & Average\\
\midrule
ChatGPT        & 75.48              & 75.00               & 75.48   & \underline{75.32}                      \\
ChatGPT-MemoryBank         & 74.68             & 80.92               & 84.56  & \underline{80.05}                 \\
ChatGPT-MemoChat   & 72.32 & 77.36 & 78.96      & \underline{76.21}\\  
\midrule
ChatGPT-Rsum (Ours)  & \textbf{78.92} & \textbf{83.56} & \textbf{84.76} & \textbf{\underline{82.41}}\\  
\bottomrule
\end{tabular}}
\end{table}
\paragraph*{Pairwise Models Evaluation}
Furthermore, we randomly sample 1000 generated responses from pairwise models, i.e., ours vs. baselines, and then ask the GPT-4 to decide which response is better based on engagingness,  consistency, and coherency. The results are shown in Figure~\ref{fig:win}. Compared to the most competitive baseline (MemoryBank), our proposed method obtains a 36.3\% improvement (winning 48.2\% and only losing 11.9\%), which illustrates the advancement of the proposed iteration mechanism.
\label{sec:analysis}
\subsection{Ablation Study}
To better understand the effectiveness of proposed memory mechanisms for LLMs, we employ ChatGPT as the LLM and conduct ablation studies in session 5. The results are shown in Table~\ref{tab:abla}. First, we only use the dialogue context of the current session as the input of LLMs, named ``W/O Memory''. As expected, the performance of the model has decreased significantly, proving the necessity of available memories from the past in a long-term conversation. Second, we replace the generated memory with ground truth memory, i.e., prompt ChatGPT to generate responses using golden memory and dialogue context, named ``Gt. Memory''. Interestingly, the model gains lower BLEU and F1 scores than using predicted memory (Ours). The potential reason is that the golden memories, e.g., ``I am trying to lose weight'' and ``I want to start running'', are fragmented and lack cohesiveness, which is sub-optimal as the LLM's prompt, whereas recursively summarizing memory generation method could wisely model the long dependencies, and generate the easy-to-digest prompt for LLMs. More analysis can be seen in \S\ref{sec:ana_mem} and \S\ref{sec:ana_resp}.
\begin{table}[t]
\centering
\caption{\textbf{The ablation study on memory} in MSC dataset.}
\label{tab:abla}
\resizebox{0.5\linewidth}{!}{
\begin{tabular}{lllll}
\toprule
Method       & BScore & F1    & BLEU-1/2    \\
\midrule
ChatGPT-Rsum (Ours)                  & \textbf{86.89}  & \textbf{20.48} & \textbf{21.83/12.59} \\
\midrule
W/O Memory            & 85.40            & 18.94          & 21.10/12.17          \\
Gt. Memory & 85.93           & 20.46          & 21.50/12.40   \\ \bottomrule
\end{tabular}}

\end{table}

\subsection{Analysis}
\label{sec:ana_mem}
Beyond the main results and ablation study, we also aim to delve deeper into our method. In the following part, we would like to discuss several research questions (\textbf{RQs}): \textit{\textbf{RQ1}: What is the quality of the generated memory?} \textit{\textbf{RQ2}: What errors may occur in memory generation?} \textit{\textbf{RQ3}: Is the proposed method robust to other LLMs?} \textit{\textbf{RQ4}: Can our zero-shot method be effectively applied in few-shot scenarios?}
  
\begin{figure}[t]
  \centering
  \includegraphics[scale=0.4] {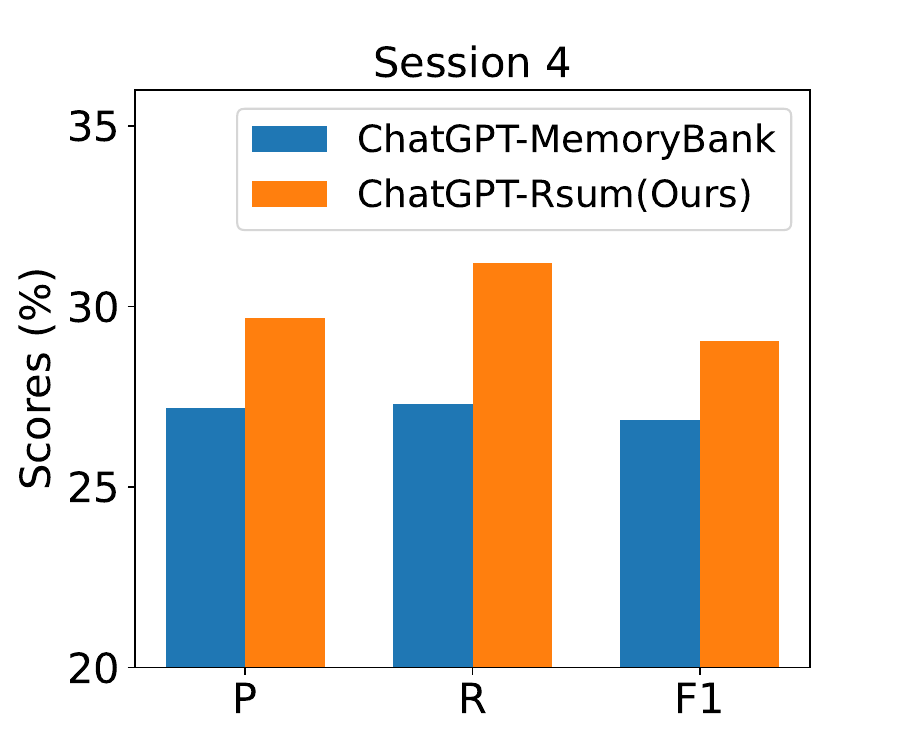}
\includegraphics[scale=0.4] {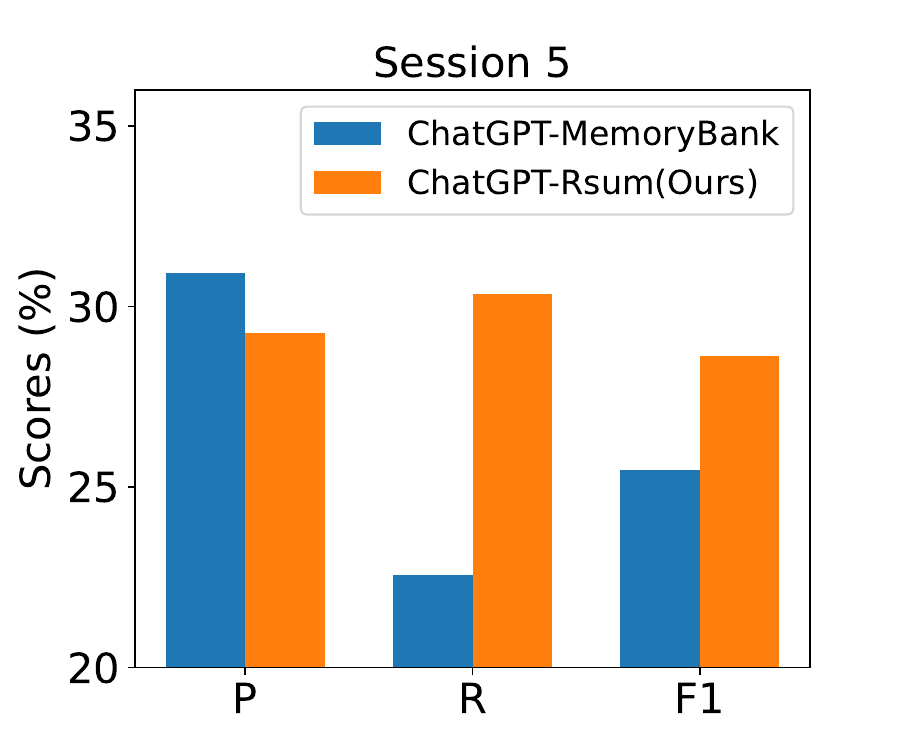}
  \caption{\textbf{The evaluation on generated memory} on ChatGPT-MemoryBank and ours. The ``P'' and ``R'' refer the precision and recall, respectively.}
  \label{fig:per_memory}
  \end{figure}

 \textbf{Our method can produce accurate and available memory (Q1).}
Central to this framework is to generate the dialog memory by continuously summarizing. To verify the quality of summarization, we compute automatic metrics between predicted memory and golden memory in the MSC dataset on ChatGPT-MemoryBank and ours, respectively, which is shown in Figure~\ref{fig:per_memory}. As seen, the generated memories of both models gain considerable F1 scores (+25\%), explaining the reliability of using dialogue summaries as memory. Besides, ChatGPT-Rsum (Ours) achieves higher overall performances on memory, suggesting that recursively summarizing can obtain more complete and long-term information than ChatGPT-MemoryBank.
Finally, given the response performances of session 5 in Table~\ref{tab:msc} (Ours \textgreater ChatGPT-MemoryBank), we suppose that the accuracy of memory prediction is positively correlated with the quality of response. We also believe advanced memory mechanisms can achieve further improvement, which might be investigated in future works.

\textbf{Our memory suffers from a few fact errors within an acceptable range (Q2).}  
 
\begin{table*}[t]
 \centering
 \caption{\textbf{Three error types in generated memory}, including corresponding examples and error proportion of content. The error context is marked in \textcolor{red}{red}.}
 \label{tab:error}
\scalebox{0.53}{
\begin{tabular}{lllll}
\toprule
\textbf{Error Type}                                                                 & \textbf{Past Dialogs}                                                                                                                                                                     & \textbf{Generated Memory}                                                                                              & \textbf{Golden Memory}                                                                                                          & \textbf{Prop.} \\
\midrule
\begin{tabular}[c]{@{}l@{}} 
Fabricated \\ Facts\end{tabular}     & \begin{tabular}[c]{@{}l@{}}Bot: I like to walk to work instead of \\ driving so I see animals.\end{tabular}                                                                              & \begin{tabular}[c]{@{}l@{}}The bot enjoys walking to work \\ \textcolor{red}{to see animals.}\end{tabular}                              & bot: I walk to work.                                                                                                         & 2.7\%                 \\
\midrule
\begin{tabular}[c]{@{}l@{}}Incorrect \\ Relationships \end{tabular} & \begin{tabular}[c]{@{}l@{}}User: I ended up giving in and getting my \\ daughters the cat.....Well when you have daughters \\ you sort of give in to them. She named it Angie.\end{tabular} & \begin{tabular}[c]{@{}l@{}}The user's daughters now have a \\cat named Angie, \textcolor{red}{which the user} \\\textcolor{red}{gave in to.}\end{tabular} & \begin{tabular}[c]{@{}l@{}}User: I got my daughters a cat.\\ My cat is named Angie. \\ My daughters named the cat.\end{tabular} & 3.2\%                 \\
\midrule
\begin{tabular}[c]{@{}l@{}}Missing \\ Details\end{tabular}                 & \begin{tabular}[c]{@{}l@{}}User: I just saw the best movie on netflix. \\ Bot: What movie did you see? \\ User: It's a documentary called The Social Dilemma.\end{tabular}               & \begin{tabular}[c]{@{}l@{}}User: Enjoys watching TV, reading, \\and listening to music\end{tabular}                   & \begin{tabular}[c]{@{}l@{}}User: I think the documentary \\ The Social Dilemma is the best \\ movie on Netflix.\end{tabular}    & 3.9\%  \\
\bottomrule
\end{tabular}}
\end{table*}
One may doubt that the generated summaries might have serious factual inconsistency and error propagation issues. We argue that summarization is not a difficult task and some works find that LLM summaries exhibit better factual consistency and fewer hallucinations\citep{Pu2023SummarizationI}. To further address this concern, we randomly choose 100 dialog samples and manually evaluate the quality of memory in the last session. Table~\ref{tab:error} reports three error types found in our generated memory. 1) \textit{Fabricated facts} refers to the memory containing some information that the dialog history cannot verify. In the first case, the bot walks to work not to see animals. 2) \textit{Incorrect relationships} refers that the generated memory concludes error causal or references from history. In the second dialogue, the user gave in to her daughter, rather than that cat. 3) \textit{Missing details} refers that the memory drops partial details of events. In the third scenario, the model ignores the user's favorite movie name and only roughly summarizes it as enjoying watching TV.
Although there exist some mistakes in our generated memory, the incorrect/ inaccurate information does not exceed 10\% of the summaries content, indicating that most recursively generated summaries are trustworthy and usable to aid long-term dialog. 
Besides, extensive experiments (in Table~\ref{tab:msc} and Figure~\ref{fig:per_memory}) validate the efficacy of our approach in utilizing generated summaries for long-term dialogs, which constitutes the primary contribution of this paper.
Lastly, the above analysis also illustrates that our method needs to be strengthened in memorizing accurate dialog details. More advanced agent-based techniques, such as retrieval-enhanced methods~\citep{Zhang2023RetrieveAT}, can be applied in the future.
 
\begin{figure}[t]
 \centering
 \begin{minipage}[t]{0.46\textwidth}
 \centering
 \includegraphics[scale=0.4]{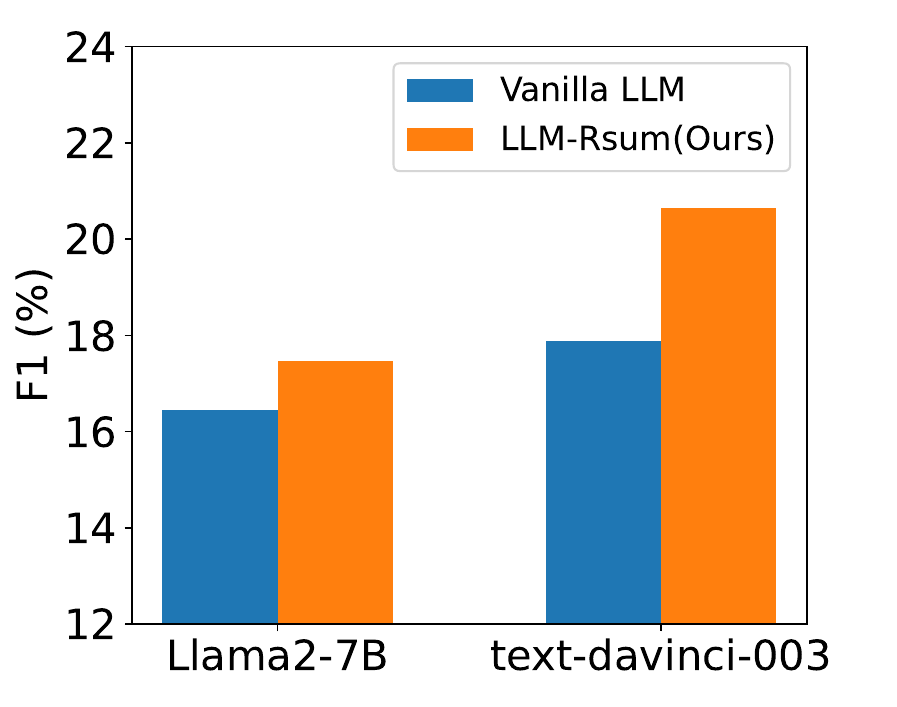}
 \caption{\textbf{The F1 score on responses} when using other LLMs.}
 \label{fig:robust_llm}
 \end{minipage}
 \hspace{0.05cm}
 \begin{minipage}[t]{0.46\textwidth}
 \centering
 \includegraphics[scale=0.4]{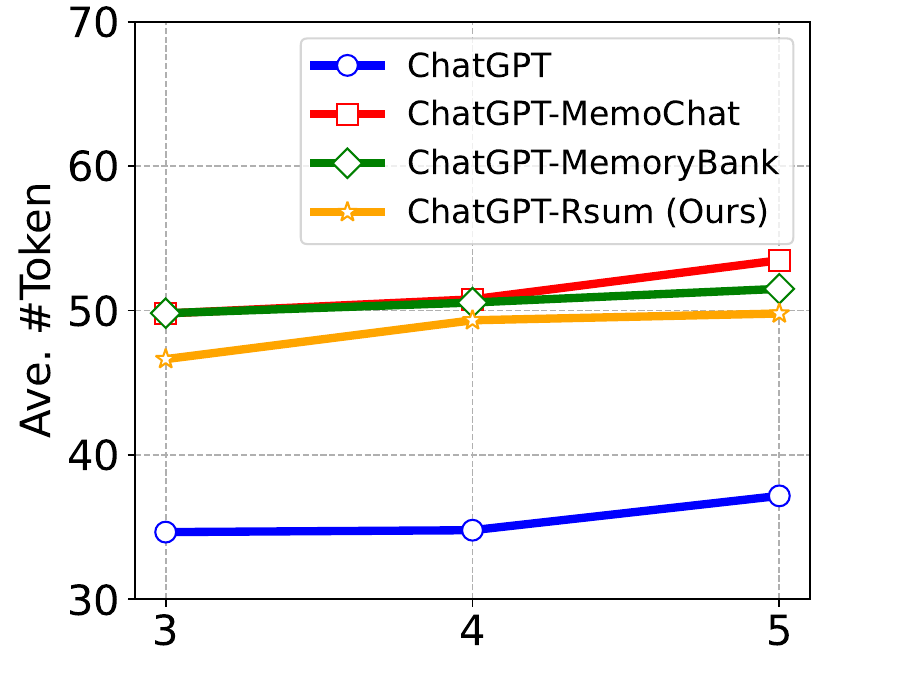}
 \caption{\textbf{The average number of tokens} 
 in generated responses.}
 \label{fig:ave_tokens}
 \end{minipage}
 \end{figure}

\textbf{Our plug-and-play method is effective for other small-scale and large-scale LLMs (Q3).} 
To check whether the proposed recursively summarizing method is robust to other large language models to tackle long-term sessions, we evaluate the framework by employing  ``Llama2-7B'' and ``text-davinci-003'' as the backbone models. 
The performances in long-context dialogue (session 5 of the MSC dataset) are shown in Figure~\ref{fig:robust_llm}, where the significant improvements confirm the robustness of our method upon different LLMs. We also believe that more powerful language models, i.e., ``text-davinci-003'' \textgreater ``Llama2-7b'', indeed understand the context better, and generate more accurate memory and responses, thus leading to better improvements from the proposed mechanism.

\textbf{Our method can be further enhanced by several labeled dialogs (Q4).}
We evaluate the few-shot performance of the proposed mechanism by the in-context technique. In precise, we utilize several dialogues with generated memory and labeled responses (ground truth), randomly sampled from the valid set, to prompt the response generator before the test inputs.
Table~\ref{tab:few_shot} shows that even two labeled samples can bring obvious advantages under our framework, on both F1 and BLEU scores, indicating the potential of our framework. We analyze that the generated memory may contain a significant amount of speaker preference information, which undoubtedly increases the difficulty of generating replies. Therefore, labeled data is highly valuable for the proposed method, as it naturally guides the LLM in utilizing the memory effectively.
\begin{table}[t]
\centering
\caption{The comparative \textbf{results (\%) on zero-shot and few-shot when using generated memory} in MSC dataset.}
\label{tab:few_shot}
\resizebox{0.5\linewidth}{!}{
\begin{tabular}{lllll}
\toprule
{\textbf{N-shot}} & \multicolumn{2}{c}{\textbf{Session 4}} & \multicolumn{2}{c}{\textbf{Session 5}}  \\ 
\cmidrule(lr){2-3} \cmidrule(lr){4-5}
&F1         & BLEU-1/2         & F1         & BLEU-1/2         \\ \midrule
Zero-shot          & 20.19        & 21.80/12.57       & 20.48        & 21.76/12.59       \\
Two-shot           & 20.37        & 22.11/12.65       & 20.63        & 22.04/12.71       \\ 
Three-shot           & 20.98        & 22.43/12.76       & 21.08        & 22.23/12.82       \\
\bottomrule
\end{tabular}}
\end{table}

\begin{figure*}[t]
  \centering
  \resizebox{\linewidth}{!}{
  \includegraphics[scale=1.1]{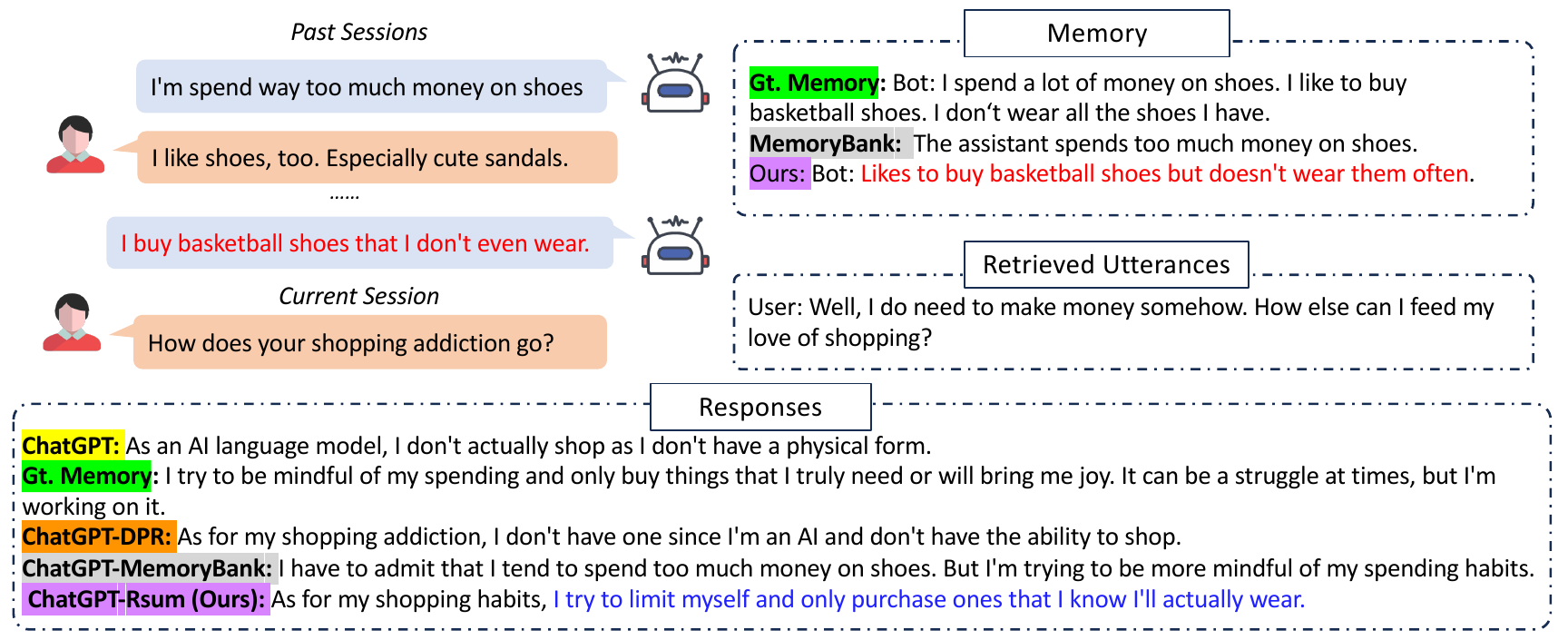}}
  \caption{\textbf{Generated responses when using different methods} in MSC dataset. Among them, our framework can obtain the \textcolor{red}{up-to-date memory} and incorporate it into \textcolor{blue}{generated response}. For clarity, we omit other utterances and predicted memory unrelated to the current query.}
  \label{fig:case}
\end{figure*}
\subsection{Case Study}
\label{sec:ana_resp}
To check whether the memory incorporates long-range dialogue information into the response, we first compare the length of the response when using different methods. As shown in Figure~\ref{fig:ave_tokens}, the average response length of using generated memory is about 15 tokens longer than the vanilla LLM (without memory) across all sessions. Furthermore, we take a sample from the generated responses to analyze the impact of memory on different methods. In the case of Figure~\ref{fig:case}, the user mentions \textit{``shopping addiction''} at the current turn, referring to the bot's habit of buying too many shoes. From the result, we can draw the following conclusions: (1) The retriever-based method (ChatGPT-DPR) and context-only LLM (ChatGPT) tend to focus on local (or closest) dialog information with long-context inputs. (2) Compared to using the golden memory, the generated memory is more fluent and coherent. It also explains why our method outperforms better than using golden memory directly, which has been observed in Table~\ref{tab:abla}. (3) Compared to the competitive memory mechanism (MemoryBank), our method can iterate and update the memory promptly, keeping consistent with the ongoing conversation.
(4) The proposed recursively summarized memory method indeed integrates long-term dialogue information into generated responses. In Figure~\ref{fig:case}, the latest memory (i.e., the bot's preference for basketball shoes) is understood and mentioned in the response.
\subsection{Complementary to Existing Works}
Our proposed method is a new memory mechanism for improving the long-range dialogue ability of LLMs, which is expected to complement existing works, including retrieval-based and input-expended methods. Here, we list two representative approaches and show the orthogonality.
\paragraph{Retrieval-based LLMs.} The effectiveness of retrieval-based methods on the MSC dataset can be observed in Table~\ref{tab:msc}, illustrating their potential in long-range conversations. Here, we further explore the complementary to the proposed method. As shown in Table~\ref{tab:retrieval}, retrieval-enhanced methods (ChatGPT-BM25 and ChatGPT-DPR) achieve further improvements than vanilla ChatGPT, showing the importance of recalling relevant information in long-range dialogs. Besides, using our framework could push these retrieval-based methods toward better performance, gaining about +0.8\% of F1 scores. We explain that these retrieved utterances can be viewed as evidence of event details, which together with our generated memory enhances LLM's long-term dialog ability.
\begin{table}[t]
\centering
\caption{\textbf{Complementarity between ours and retrieval-based methods}, in terms of automatic and LLM's evaluation on session 5 of the MSC dataset.}
\label{tab:retrieval}
\resizebox{0.5\linewidth}{!}{
\begin{tabular}{llll}
\toprule
\textbf{Method}                                            & F1    & Coherence &Consistency \\ \midrule
ChatGPT &19.41 & 75.00 & 75.48\\ \midrule
ChatGPT-BM25 (k=5)  & 20.91 & 75.44          & 76.88            \\
+ Our framework  & \textbf{21.81}           &    \textbf{84.44}                &   \textbf{90.68}                   \\
\midrule
ChatGPT-DPR (k=5) & 20.97          & 78.60               & 79.20                 \\
+ Our framework  &       \textbf{21.69}           &   \textbf{83.40}                 & \textbf{86.48}                     \\ \bottomrule
\end{tabular}}

\end{table}
\begin{table}[t]
\centering
\caption{\textbf{Complementarity between ours and long-context LLMs}, in terms of automatic and LLM's evaluation on session 5 of the MSC dataset.}
\label{tab:long_context}
\resizebox{0.6\linewidth}{!}{
\begin{tabular}{llll}
\toprule
\textbf{Method}                                            & F1    & Coherence & Consistency \\ \midrule
longlora-8k                        & 14.02 & 42.44          & 62.04            \\
longlora-8k + Our framework      & \textbf{15.77}          &   \textbf{53.41}                 &      \textbf{68.60}                \\ \midrule
ChatGPT & 19.41 & 75.00 & 75.48\\ 
ChatGPT-16k & \textbf{19.92}          & 78.60               & 79.20                 \\
ChatGPT-16k+ Our framework                                          & 19.29               &    \textbf{90.04}                &  \textbf{92.44}                    \\ \midrule
GPT-4o                      & 20.35 & 87.70          & 82.00           \\
GPT-4o + Our framework      & \textbf{21.02}      &   \textbf{91.12}                 &      \textbf{93.29}             \\
\bottomrule
\end{tabular}}
\end{table}

 \paragraph{Long-context LLMs.}  To process the entire context and reduce the information loss, many researchers try to extend the context length of LLMs via training from scratch, fine-tuning, or other specific algorithms (e.g., FlashAttention)~\citep{Dao2022FlashAttentionFA}. For example, LongLoRA~\citep{Chen2023LongLoRAEF} extends Llama2~\citep{llama2} 7B from 4k context to 100k. Although the maximum length of the datasets used is no more than 4k, which accommodates most popular LLMs, we still want to explore whether the proposed method is complementary to these length-extended models. Here, we utilize three popular LLMs with large context windows, i.e., ``longlora-8k''\footnote{\url{https://huggingface.co/Yukang/Llama-2-7b-longlora-8k}},  ChatGPT-16k (the version ``gpt-3.5-turbo-16k-0613'') and GPT-4o\footnote{\url{https://openai.com/index/hello-gpt-4o/}}to verify the effectiveness of our proposed framework. To ensure the quality of generated memory, we use the ChatGPT ``gpt-3.5-turbo-0301'' as the memory iterator, and only apply the above long-context LLMs to finish the memory-based response generation. From the results in Table~\ref{tab:long_context}, we conclude that: 1) Increasing the maximum context length of LLMs indeed enhances the long-term ability of dialog, with obvious improvements in coherency and consistency, even when the input length is much smaller than the context window. 2) Even if the input length does not exceed the window size, applying our method to LLMs remains an effective way to recall information and maintain long-term connections. 3) With a stronger LLM as the backbone (e.g., GPT-4o > ChatGPT > longloRA), the model achieves better performance after our enhancements. We explain that our recursive summaries help reorganize and digest past information efficiently, thereby enhancing the understanding of semantics in long-range conversations.

\section{Conclusion}
In this paper, we propose a simple and effective strategy by recursively summarizing to improve the long-term dialogue ability in LLMs. The experimental results indicate our generated memory can model long-term dependencies and prompt LLMs to generate highly consistent responses. 
Additional analysis indicates that the method is robust across different LLMs and can be further boosted in few-shot scenarios. Importantly, our method also shows strong complementary to both popular retrieval-based and long-context models.

\paragraph{Limitations}
One of the primary limitations of our method is that it does not account for the cost associated with calling large models. This is a significant factor that cannot be overlooked in real-life applications, where computational resources and associated costs are often a constraint. Besides, while our generated memory is effective, it occasionally suffers from minor factual errors. These inaccuracies, though few, highlight an area for improvement that can be addressed in future research.

\paragraph*{Future work} One promising direction for future work is to explore the effectiveness of our method in modeling long-context tasks beyond dialogue, such as story generation. Investigating how well our approach can handle different types of long-context tasks will provide deeper insights into its versatility and potential applications. Another avenue for future research is to optimize the performance of our summarization method by using a locally supervised, fine-tuned LLM. This approach could potentially reduce the reliance on expensive online APIs, making the method more accessible and cost-effective for broader use.
\section{Acknowledgements}
This work is supported by the National Natural Science Foundation
of China (No.U2336202).

\appendix
\section{Dataset Statistics}
\label{ap:dataset}
The statics of MSC and Carecall are shown in Table~\ref{tab:msc_dataset} and Table~\ref{tab:care_dataset}. The session ID $i$ indicates there are $i-1$ history conversation sessions that happened before the last session. The ``\#Ave. Tokens'' and ``\#Max. Tokens''  refer to the average and maximum number of tokens of dialogs, respectively.
\begin{table}[htbp]
\centering
\caption{\textbf{The statistics of MSC Dataset}.}
\label{tab:msc_dataset}
\resizebox{0.7\linewidth}{!}{
\begin{tabular}{lllll}
\toprule
\textbf{Session ID} & \textbf{\#Dialog} & \textbf{\#Response} & \textbf{\#Ave. Tokens} & \textbf{\#Max. Tokens}\\ \midrule
Session 2   & 501      & 5,939 & 444.84  &951    \\
Session 3   & 501      & 5,924 & 810.19  &1733    \\
Session 4   & 501      & 5,940 & 1195.08  &2234    \\
Session 5   & 501      & 5,945 & 1598.78 &2613   \\ \bottomrule
\end{tabular}}
\end{table}

\begin{table}[htbp]
\centering
\caption{\textbf{The statistics of Carecall Dataset}.}
\label{tab:care_dataset}
\resizebox{0.7\linewidth}{!}{
\begin{tabular}{lllll}
\toprule
\textbf{Session ID} & \textbf{\#Dialog} & \textbf{\#Response} & \textbf{\#Ave. Tokens} & \textbf{\#Max. Tokens}\\ \midrule
Session 2   & 2798      & 7826 & 285.03  &692   \\
Session 3   & 743      & 7693 & 459.63  &1093    \\
Session 4   & 674      & 7065 & 636.55  &1418    \\
Session 5   & 638      & 6553 & 809.45 &1744   \\ \bottomrule
\end{tabular}}
\end{table}
\section{Prompt Designs}
\label{ap:prompt}
The following are all prompts utilized in our experiments:
\begin{itemize}
    \item Context-only LLMs (\textbf{Llama-7B}, \textbf{ChatGLM-6B}, and \textbf{ChatGPT}): Table~\ref{tab:context_only}
    \item Retrieval-based LLMs (\textbf{ChatGPT-BM25} \& \textbf{ChatGPT-DPR}): Table~\ref{tab:retrieval_model}
    \item Retrieval-based LLMs enhanced by our framework (\textbf{ChatGPT-BM25 + Our framework} \& \textbf{ChatGPT-DPR + Our framework}): Table~\ref{tab:retrieval_model_mem}
    \item Our method enhanced by in-context learning (Take the one-shot as an example): Table~\ref{tab:in_context_learn}
    \item LLM evaluations: Table~\ref{tab:single_model} and Table~\ref{tab:pairwise_model}
\end{itemize}

\begin{table}[htbp]
\centering
\caption{The prompt for the context-only LLM.}
\label{tab:context_only}
\scalebox{0.68}{
\begin{tabular}{ll}
\toprule
\textbf{Prompt} &  \begin{tabular}[c]{@{}l@{}}You are an advanced AI language model capable of engaging in personality-based conversations. \\ Respond to the user based on the provided dialogue context. Craft a response that is natural and \\ conversational.\\ Dialog context: {[}dialog{]}\\ The response to user is: \end{tabular} \\ \midrule
\textbf{Output} & {[}response{]}                                                                                                                                                                                                                                                                                                                                                 \\ 
\bottomrule
\end{tabular}}
\end{table}

\begin{table}[htbp]
\centering
\caption{The prompt for the retrieval-based LLM.}
\label{tab:retrieval_model}
\scalebox{0.65}{
\begin{tabular}{ll}
\toprule
\textbf{Prompt} & \begin{tabular}[c]{@{}l@{}}You are an advanced AI designed for engaging in natural, personality-based conversations. \\You will be provided with dialogue memory, relevant historical context, and dialogue context. \\The dialogue memory contains the personality, preferences, and experiences of the speakers \\(the user and the assistant). When responding, consider maintaining a conversational and fluent tone. \\Responses should be contextually relevant and aim to keep the conversation flowing. \\Relevant context: {[}retrieved utterances{]}. Dialogue context: {[}dialog{]}. \\So the response to the user is: \end{tabular} \\ 
\midrule
\textbf{Output} & {[}response{]} \\ 
\bottomrule
\end{tabular}}
\end{table}

\begin{table}[t]
\centering
\caption{The prompt for the retrieval-based LLM enhanced by our framework.}
\label{tab:retrieval_model_mem}
\scalebox{0.7}{
\begin{tabular}{ll}
\toprule
\textbf{Prompt} & \begin{tabular}[c]{@{}l@{}}You are an advanced AI designed for engaging in natural, personality-based conversations. \\You will be provided with dialogue memory, relevant historical context, and dialogue context. \\The dialog memory contains the personality, preferences and experiences of speakers  (the \\assistant and the user). When responding, consider maintaining a conversational and fluent tone. \\Responses should be contextually relevant and aim to keep the conversation flowing. \\ Relevant context: {[}retrieval utterances{]} \\ Memory: {[}latest memory{]} \\ Dialogue: {[}current context{]}\end{tabular} \\ \midrule
\textbf{Output} & {[}response{]}                             \\ \bottomrule
\end{tabular}}
\end{table}

\begin{table}[htbp]
\centering
\caption{The prompt for our method with in-context learning.}
\label{tab:in_context_learn}
\scalebox{0.73}{
\begin{tabular}{ll}
\toprule
\textbf{Prompt} &  \begin{tabular}[c]{@{}l@{}}You are an advanced AI designed for engaging in natural, personality-based conversations. \\You will be provided with a memory, containing the personal preferences and experiences \\ of speakers (the assistant and the user), as well as a dialogue context. When responding, \\consider maintaining a conversational and fluent tone. Responses should be \\contextually relevant, consistent with given memory, aiming to keep the conversation \\ flowing. Your goal is to provide engaging and coherent responses based on the \\dialogue context provided. To help you understand this task, we provide 1 example below. \\ EXAMPLE 1: The example memory is: \\ User: - Divorced - Raising one child - Immigrated from Britain last year - Metal worker\\ Assistant: - Not married - Girlfriend has 2 kids - Works on mTurk, landscaping, sales,\\ envelope stuffing, painting - Used to love winter, but has become intolerant of it\\ - Works with a friend who owns "John of all trades"\\ The example dialogue context is: \\ User: Today was the hottest day I've ever experienced here in Florida!\\ So the response to the user is: do you enjoy the heat more than the cold in Britain?\\ The following is the case you need to test: The test memory is:{[}previous memory{]} \\The test dialogue context is:{[}dialog{]} So the response to the user is:\end{tabular} \\ \midrule
\textbf{Output} & {[}response{]}                                                                                                                                                                                                                                                                                                                                                                                                                                                                                                                                                                                                                                                                                                                                                                                                                                                                                                                                                                                                                                                                                                                                                                                                                                                                                                                                                                                                                                                                               \\ \bottomrule
\end{tabular}}
\end{table}

\begin{table}[htbp]
\centering
\caption{The prompt for single model evaluation.}

\label{tab:single_model}
\scalebox{0.73}{
\begin{tabular}{ll}
\toprule
\textbf{Prompt} & \begin{tabular}[c]{@{}l@{}}You are an impartial judge. You will be shown a Conversation Context, Personality of \\Speakers and Assistant Response.\\ \#Fluency: Please evaluate whether the Assistant's response is natural, fluent, and similar to \\human communication, avoiding repetition and ensuring a diverse range of output.\\ \#Consistency: Please evaluate whether the Assistant's response is consistent with the \\information of persona list. Any deviation from the expected personality may indicate a lack \\of coherence.\\ \#Coherency: Please evaluate whether the Assistant's response maintains a coherent and logical \\flow  of conversation based on the evolving context. A response with good context coherence \\ can understand and respond appropriately to changes in conversation topics, providing smooth \\and sensible interactions.\\ Conversation Context:{[}dialog{]} Personality:{[}persona{]} Assistant Response: {[}response{]}\\ Begin your evaluation by providing a short explanation, then you must rate the Assistant \\Response on an integer score of 1 (very bad) to 100 (very good) by strictly following this format: \\ {[}{[}score{]}{]}.\end{tabular} \\ \midrule
\textbf{Output} & {[}output{]}                                                                                                                                                                                                                                                                                                                                                                                                                                                                                                                                                                                                                                                                                                                                                                                                                                                                                                                                                                                                                                                                                                                                                                                                                                                          \\ \bottomrule
\end{tabular}}
\end{table}

\begin{table}[htbp]
\centering
\caption{The prompt of pairwise model evaluation.}
\label{tab:pairwise_model}
\scalebox{0.73}{
\begin{tabular}{ll}
\toprule
\textbf{Prompt} & \begin{tabular}[c]{@{}l@{}}Hi! We are a group of researchers working on Artificial Intelligence. In this task, we will ask \\you to help us rate the assistant's responses. In the area below, you will first read: \\ 1. A conversation context comes from two speakers (the user and bot) \\ 2. The personality of two speakers (the user and bot) extracted from past dialogs. \\ 3. Two responses from AI systems. Your task is to decide which response is better. There are \\ several dimensions that you can think along. Consider the following questions:\\ 1. Is the response coherent? A response with good context coherence can understand and \\respond appropriately to changes in conversation topics,  providing smooth and sensible interactions.\\ 2. Is the response consistent? Evaluate whether the response \\ is consistent with the information of persona list. Any  deviation from the expected personality may\\ indicate a lack of consistency.\\ 3. Is the response natural and fluent? Please evaluate whether the response is natural, fluent, \\and similar to human communication,  avoiding excessive repetition and ensuring a  diverse range \\of output. \\ Based on your aesthetics, which one do you prefer? For example, you might prefer one poem over \\another poem. Ultimately, you should decide which response is better based on your judgment and\\ your own preference. There are four options for you to choose from: \\ 1.Response 1 is better : If you think response 1 has an advantage, then choose this option. \\ 2.Response 1 is slightly better : Response 1 is very marginally better than response 2 and the \\difference is small. \\ 3.Response 2 is slightly better : Response 2 is very marginally  better than response 1 and the \\difference is small. \\ 4.Response 2 is better : If you think response 2 has an advantage, then choose this option. \\ There are cases where the difference between the two responses is not clear. In this case, \\you can choose the second or the third option. However, in general, we ask you to choose \\ those options as few as possible.\\ response 1: {[}response1{]} response 2: {[}response2{]}\end{tabular} \\ \midrule
\textbf{Output} & {[}response{]}                                                                                   \\ \bottomrule
\end{tabular}}
\end{table}

\clearpage
\bibliographystyle{elsarticle-harv} 
\bibliography{ref}

\begin{thebibliography}{47}
\expandafter\ifx\csname natexlab\endcsname\relax\def\natexlab#1{#1}\fi
\providecommand{\url}[1]{\texttt{#1}}
\providecommand{\href}[2]{#2}
\providecommand{\path}[1]{#1}
\providecommand{\DOIprefix}{doi:}
\providecommand{\ArXivprefix}{arXiv:}
\providecommand{\URLprefix}{URL: }
\providecommand{\Pubmedprefix}{pmid:}
\providecommand{\doi}[1]{\href{http://dx.doi.org/#1}{\path{#1}}}
\providecommand{\Pubmed}[1]{\href{pmid:#1}{\path{#1}}}
\providecommand{\bibinfo}[2]{#2}
\ifx\xfnm\relax \def\xfnm[#1]{\unskip,\space#1}\fi
\bibitem[{Achiam et~al.(2023)Achiam, Adler, Agarwal, Ahmad, Akkaya, Aleman, Almeida, Altenschmidt, Altman, Anadkat et~al.}]{GPT4OpenAI}
\bibinfo{author}{Achiam, J.}, \bibinfo{author}{Adler, S.}, \bibinfo{author}{Agarwal, S.}, \bibinfo{author}{Ahmad, L.}, \bibinfo{author}{Akkaya, I.}, \bibinfo{author}{Aleman, F.L.}, \bibinfo{author}{Almeida, D.}, \bibinfo{author}{Altenschmidt, J.}, \bibinfo{author}{Altman, S.}, \bibinfo{author}{Anadkat, S.}, et~al., \bibinfo{year}{2023}.
\newblock \bibinfo{title}{Gpt-4 technical report}.
\newblock \bibinfo{journal}{arXiv preprint arXiv:2303.08774} .
\bibitem[{An et~al.(2023)An, Gong, Zhong, Li, Zhang, Kong and Qiu}]{An2023LEvalIS}
\bibinfo{author}{An, C.}, \bibinfo{author}{Gong, S.}, \bibinfo{author}{Zhong, M.}, \bibinfo{author}{Li, M.}, \bibinfo{author}{Zhang, J.}, \bibinfo{author}{Kong, L.}, \bibinfo{author}{Qiu, X.}, \bibinfo{year}{2023}.
\newblock \bibinfo{title}{L-eval: Instituting standardized evaluation for long context language models}.
\newblock \bibinfo{journal}{arXiv preprint arXiv:2307.11088} \bibinfo{volume}{abs/2307.11088}.
\bibitem[{Bae et~al.(2022)Bae, Kwak, Kang, Lee, Kim, Jeong, Kim, Lee, Park and Sung}]{bae-etal-2022-keep}
\bibinfo{author}{Bae, S.}, \bibinfo{author}{Kwak, D.}, \bibinfo{author}{Kang, S.}, \bibinfo{author}{Lee, M.Y.}, \bibinfo{author}{Kim, S.}, \bibinfo{author}{Jeong, Y.}, \bibinfo{author}{Kim, H.}, \bibinfo{author}{Lee, S.W.}, \bibinfo{author}{Park, W.}, \bibinfo{author}{Sung, N.}, \bibinfo{year}{2022}.
\newblock \bibinfo{title}{Keep me updated! memory management in long-term conversations}, in: \bibinfo{booktitle}{Findings of the Conference on Empirical Methods in Natural Language Processing}.
\bibitem[{Beltagy et~al.(2020)Beltagy, Peters and Cohan}]{Beltagy2020LongformerTL}
\bibinfo{author}{Beltagy, I.}, \bibinfo{author}{Peters, M.E.}, \bibinfo{author}{Cohan, A.}, \bibinfo{year}{2020}.
\newblock \bibinfo{title}{Longformer: the long-document transformer}.
\newblock \bibinfo{journal}{arXiv preprint arXiv:2004.05150} .
\bibitem[{Brown et~al.(2020)Brown, Mann, Ryder, Subbiah, Kaplan, Dhariwal, Neelakantan, Shyam, Sastry, Askell, Agarwal, Herbert-Voss, Krueger, Henighan, Child, Ramesh, Ziegler, Wu, Winter, Hesse, Chen, Sigler, Litwin, Gray, Chess, Clark, Berner, McCandlish, Radford, Sutskever and Amodei}]{Brown2020LanguageMA}
\bibinfo{author}{Brown, T.B.}, \bibinfo{author}{Mann, B.}, \bibinfo{author}{Ryder, N.}, \bibinfo{author}{Subbiah, M.}, \bibinfo{author}{Kaplan, J.}, \bibinfo{author}{Dhariwal, P.}, \bibinfo{author}{Neelakantan, A.}, \bibinfo{author}{Shyam, P.}, \bibinfo{author}{Sastry, G.}, \bibinfo{author}{Askell, A.}, \bibinfo{author}{Agarwal, S.}, \bibinfo{author}{Herbert-Voss, A.}, \bibinfo{author}{Krueger, G.}, \bibinfo{author}{Henighan, T.J.}, \bibinfo{author}{Child, R.}, \bibinfo{author}{Ramesh, A.}, \bibinfo{author}{Ziegler, D.M.}, \bibinfo{author}{Wu, J.}, \bibinfo{author}{Winter, C.}, \bibinfo{author}{Hesse, C.}, \bibinfo{author}{Chen, M.}, \bibinfo{author}{Sigler, E.}, \bibinfo{author}{Litwin, M.}, \bibinfo{author}{Gray, S.}, \bibinfo{author}{Chess, B.}, \bibinfo{author}{Clark, J.}, \bibinfo{author}{Berner, C.}, \bibinfo{author}{McCandlish, S.}, \bibinfo{author}{Radford, A.}, \bibinfo{author}{Sutskever, I.}, \bibinfo{author}{Amodei, D.}, \bibinfo{year}{2020}.
\newblock \bibinfo{title}{Language models are few-shot learners}.
\newblock \bibinfo{journal}{arXiv preprint ArXiv:2005.14165} .
\bibitem[{Chen et~al.(2024a)Chen, Li, Huang, Wang and Li}]{chen2024compress}
\bibinfo{author}{Chen, N.}, \bibinfo{author}{Li, H.}, \bibinfo{author}{Huang, J.}, \bibinfo{author}{Wang, B.}, \bibinfo{author}{Li, J.}, \bibinfo{year}{2024}a.
\newblock \bibinfo{title}{Compress to impress: Unleashing the potential of compressive memory in real-world long-term conversations}.
\newblock \bibinfo{journal}{arXiv preprint arXiv:2402.11975} .
\bibitem[{Chen et~al.(2023)Chen, Wong, Chen and Tian}]{Chen2023ExtendingCW}
\bibinfo{author}{Chen, S.}, \bibinfo{author}{Wong, S.}, \bibinfo{author}{Chen, L.}, \bibinfo{author}{Tian, Y.}, \bibinfo{year}{2023}.
\newblock \bibinfo{title}{Extending context window of large language models via positional interpolation}.
\newblock \bibinfo{journal}{arXiv preprint arXiv:2306.15595} .
\bibitem[{Chen et~al.(2024b)Chen, Qian, Tang, Lai, Liu, Han and Jia}]{Chen2023LongLoRAEF}
\bibinfo{author}{Chen, Y.}, \bibinfo{author}{Qian, S.}, \bibinfo{author}{Tang, H.}, \bibinfo{author}{Lai, X.}, \bibinfo{author}{Liu, Z.}, \bibinfo{author}{Han, S.}, \bibinfo{author}{Jia, J.}, \bibinfo{year}{2024}b.
\newblock \bibinfo{title}{Longlora: Efficient fine-tuning of long-context large language models}, in: \bibinfo{booktitle}{the International Conference on Learning Representations}.
\bibitem[{Choi et~al.(2023)Choi, On, Han, Kim, Nam, Jo, Rho, Kwon and Seo}]{choi2023effortless}
\bibinfo{author}{Choi, E.}, \bibinfo{author}{On, K.W.}, \bibinfo{author}{Han, G.}, \bibinfo{author}{Kim, S.}, \bibinfo{author}{Nam, D.W.}, \bibinfo{author}{Jo, D.}, \bibinfo{author}{Rho, S.E.}, \bibinfo{author}{Kwon, T.}, \bibinfo{author}{Seo, M.}, \bibinfo{year}{2023}.
\newblock \bibinfo{title}{Effortless integration of memory management into open-domain conversation systems}.
\newblock \bibinfo{journal}{arXiv preprint arXiv:2305.13973} .
\bibitem[{Dao et~al.(2022)Dao, Fu, Ermon, Rudra and R{\'e}}]{Dao2022FlashAttentionFA}
\bibinfo{author}{Dao, T.}, \bibinfo{author}{Fu, D.}, \bibinfo{author}{Ermon, S.}, \bibinfo{author}{Rudra, A.}, \bibinfo{author}{R{\'e}, C.}, \bibinfo{year}{2022}.
\newblock \bibinfo{title}{Flashattention: Fast and memory-efficient exact attention with io-awareness}.
\newblock \bibinfo{journal}{Advances in Neural Information Processing Systems} \bibinfo{volume}{35}, \bibinfo{pages}{16344--16359}.
\bibitem[{Deriu et~al.(2019)Deriu, Rodrigo, Otegi, Echegoyen, Rosset, Agirre and Cieliebak}]{Deriu2019SurveyOE}
\bibinfo{author}{Deriu, J.}, \bibinfo{author}{Rodrigo, {\'A}.}, \bibinfo{author}{Otegi, A.}, \bibinfo{author}{Echegoyen, G.}, \bibinfo{author}{Rosset, S.}, \bibinfo{author}{Agirre, E.}, \bibinfo{author}{Cieliebak, M.}, \bibinfo{year}{2019}.
\newblock \bibinfo{title}{Survey on evaluation methods for dialogue systems}.
\newblock \bibinfo{journal}{Artificial Intelligence Review} \bibinfo{volume}{54}, \bibinfo{pages}{755 -- 810}.
\bibitem[{Dubois et~al.(2024)Dubois, Li, Taori, Zhang, Gulrajani, Ba, Guestrin, Liang and Hashimoto}]{Dubois2023AlpacaFarmAS}
\bibinfo{author}{Dubois, Y.}, \bibinfo{author}{Li, C.X.}, \bibinfo{author}{Taori, R.}, \bibinfo{author}{Zhang, T.}, \bibinfo{author}{Gulrajani, I.}, \bibinfo{author}{Ba, J.}, \bibinfo{author}{Guestrin, C.}, \bibinfo{author}{Liang, P.S.}, \bibinfo{author}{Hashimoto, T.B.}, \bibinfo{year}{2024}.
\newblock \bibinfo{title}{Alpacafarm: A simulation framework for methods that learn from human feedback}.
\newblock \bibinfo{journal}{Advances in Neural Information Processing Systems} \bibinfo{volume}{36}.
\bibitem[{GLM et~al.(2024)GLM, Zeng, Xu, Wang, Zhang, Yin, Rojas, Feng, Zhao, Lai et~al.}]{glm2024chatglm}
\bibinfo{author}{GLM, T.}, \bibinfo{author}{Zeng, A.}, \bibinfo{author}{Xu, B.}, \bibinfo{author}{Wang, B.}, \bibinfo{author}{Zhang, C.}, \bibinfo{author}{Yin, D.}, \bibinfo{author}{Rojas, D.}, \bibinfo{author}{Feng, G.}, \bibinfo{author}{Zhao, H.}, \bibinfo{author}{Lai, H.}, et~al., \bibinfo{year}{2024}.
\newblock \bibinfo{title}{Chatglm: A family of large language models from glm-130b to glm-4 all tools}.
\newblock \bibinfo{journal}{arXiv preprint arXiv:2406.12793} .
\bibitem[{Guu et~al.(2020)Guu, Lee, Tung, Pasupat and Chang}]{guu2020retrieval}
\bibinfo{author}{Guu, K.}, \bibinfo{author}{Lee, K.}, \bibinfo{author}{Tung, Z.}, \bibinfo{author}{Pasupat, P.}, \bibinfo{author}{Chang, M.}, \bibinfo{year}{2020}.
\newblock \bibinfo{title}{Retrieval augmented language model pre-training}, in: \bibinfo{booktitle}{the International Conference on Learning Representations}.
\bibitem[{Kann et~al.(2022)Kann, Ebrahimi, Koh, Dudy and Roncone}]{Kann2022OpendomainDG}
\bibinfo{author}{Kann, K.}, \bibinfo{author}{Ebrahimi, A.}, \bibinfo{author}{Koh, J.J.}, \bibinfo{author}{Dudy, S.}, \bibinfo{author}{Roncone, A.}, \bibinfo{year}{2022}.
\newblock \bibinfo{title}{Open-domain dialogue generation: What we can do, cannot do, and should do next}, in: \bibinfo{booktitle}{NLP4CONVAI}.
\bibitem[{Karpukhin et~al.(2020)Karpukhin, Oguz, Min, Lewis, Wu, Edunov, Chen and Yih}]{karpukhin-etal-2020-dense}
\bibinfo{author}{Karpukhin, V.}, \bibinfo{author}{Oguz, B.}, \bibinfo{author}{Min, S.}, \bibinfo{author}{Lewis, P.}, \bibinfo{author}{Wu, L.}, \bibinfo{author}{Edunov, S.}, \bibinfo{author}{Chen, D.}, \bibinfo{author}{Yih, W.t.}, \bibinfo{year}{2020}.
\newblock \bibinfo{title}{Dense passage retrieval for open-domain question answering}, in: \bibinfo{booktitle}{the Conference on Empirical Methods in Natural Language Processing}.
\bibitem[{Lee et~al.(2023)Lee, Hartmann, Park, Papailiopoulos and Lee}]{Lee2023PromptedLA}
\bibinfo{author}{Lee, G.}, \bibinfo{author}{Hartmann, V.}, \bibinfo{author}{Park, J.}, \bibinfo{author}{Papailiopoulos, D.}, \bibinfo{author}{Lee, K.}, \bibinfo{year}{2023}.
\newblock \bibinfo{title}{Prompted llms as chatbot modules for long open-domain conversation}, in: \bibinfo{booktitle}{Findings of the Annual Meeting of the Association for Computational Linguistics}.
\bibitem[{Lewis et~al.(2020)Lewis, Perez, Piktus, Petroni, Karpukhin, Goyal, Kuttler, Lewis, tau Yih, Rockt{\"a}schel, Riedel and Kiela}]{Lewis2020RetrievalAugmentedGF}
\bibinfo{author}{Lewis, P.}, \bibinfo{author}{Perez, E.}, \bibinfo{author}{Piktus, A.}, \bibinfo{author}{Petroni, F.}, \bibinfo{author}{Karpukhin, V.}, \bibinfo{author}{Goyal, N.}, \bibinfo{author}{Kuttler, H.}, \bibinfo{author}{Lewis, M.}, \bibinfo{author}{tau Yih, W.}, \bibinfo{author}{Rockt{\"a}schel, T.}, \bibinfo{author}{Riedel, S.}, \bibinfo{author}{Kiela, D.}, \bibinfo{year}{2020}.
\newblock \bibinfo{title}{Retrieval-augmented generation for knowledge-intensive nlp tasks}, in: \bibinfo{booktitle}{the Annual Conference on Neural Information Processing Systems}.
\bibitem[{Li et~al.(2016)Li, Galley, Brockett, Gao and Dolan}]{li-etal-2016-diversity}
\bibinfo{author}{Li, J.}, \bibinfo{author}{Galley, M.}, \bibinfo{author}{Brockett, C.}, \bibinfo{author}{Gao, J.}, \bibinfo{author}{Dolan, B.}, \bibinfo{year}{2016}.
\newblock \bibinfo{title}{A diversity-promoting objective function for neural conversation models}, in: \bibinfo{booktitle}{Annual Conference of the North American Chapter of the Association for Computational Linguistics}.
\bibitem[{Li et~al.(2023)Li, Wang, Zheng and Zhang}]{Li2023LooGLECL}
\bibinfo{author}{Li, J.}, \bibinfo{author}{Wang, M.}, \bibinfo{author}{Zheng, Z.}, \bibinfo{author}{Zhang, M.}, \bibinfo{year}{2023}.
\newblock \bibinfo{title}{Loogle: Can long-context language models understand long contexts?}
\newblock \bibinfo{journal}{arXiv preprint arXiv:2311.04939} .
\bibitem[{Liu et~al.(2016)Liu, Lowe, Serban, Noseworthy, Charlin and Pineau}]{liu-etal-2016-evaluate}
\bibinfo{author}{Liu, C.W.}, \bibinfo{author}{Lowe, R.}, \bibinfo{author}{Serban, I.}, \bibinfo{author}{Noseworthy, M.}, \bibinfo{author}{Charlin, L.}, \bibinfo{author}{Pineau, J.}, \bibinfo{year}{2016}.
\newblock \bibinfo{title}{How {NOT} to evaluate your dialogue system: An empirical study of unsupervised evaluation metrics for dialogue response generation}, in: \bibinfo{booktitle}{the Conference on Empirical Methods in Natural Language Processing}.
\bibitem[{Liu et~al.(2023)Liu, Lin, Hewitt, Paranjape, Bevilacqua, Petroni and Liang}]{Liu2023LostIT}
\bibinfo{author}{Liu, N.F.}, \bibinfo{author}{Lin, K.}, \bibinfo{author}{Hewitt, J.}, \bibinfo{author}{Paranjape, A.}, \bibinfo{author}{Bevilacqua, M.}, \bibinfo{author}{Petroni, F.}, \bibinfo{author}{Liang, P.}, \bibinfo{year}{2023}.
\newblock \bibinfo{title}{Lost in the middle: How language models use long contexts}, in: \bibinfo{booktitle}{Transactions of the Association for Computational Linguistics}.
\bibitem[{Lu et~al.(2023a)Lu, An, Lin, Pergola, He, Yin, Sun and Wu}]{Lu2023MemoChatTL}
\bibinfo{author}{Lu, J.}, \bibinfo{author}{An, S.}, \bibinfo{author}{Lin, M.}, \bibinfo{author}{Pergola, G.}, \bibinfo{author}{He, Y.}, \bibinfo{author}{Yin, D.}, \bibinfo{author}{Sun, X.}, \bibinfo{author}{Wu, Y.}, \bibinfo{year}{2023}a.
\newblock \bibinfo{title}{Memochat: Tuning llms to use memos for consistent long-range open-domain conversation}.
\newblock \bibinfo{journal}{arXiv preprint arXiv:2308.08239} .
\bibitem[{Lu et~al.(2023b)Lu, Qiu, Ding, Xie and Tao}]{Lu2023EAPrompt}
\bibinfo{author}{Lu, Q.}, \bibinfo{author}{Qiu, B.}, \bibinfo{author}{Ding, L.}, \bibinfo{author}{Xie, L.}, \bibinfo{author}{Tao, D.}, \bibinfo{year}{2023}b.
\newblock \bibinfo{title}{Error analysis prompting enables human-like translation evaluation in large language models: A case study on chatgpt}.
\newblock \bibinfo{journal}{arXiv preprint} .
\bibitem[{Mazar{\'e} et~al.(2018)Mazar{\'e}, Humeau, Raison and Bordes}]{mazare-etal-2018-training}
\bibinfo{author}{Mazar{\'e}, P.E.}, \bibinfo{author}{Humeau, S.}, \bibinfo{author}{Raison, M.}, \bibinfo{author}{Bordes, A.}, \bibinfo{year}{2018}.
\newblock \bibinfo{title}{Training millions of personalized dialogue agents}, in: \bibinfo{booktitle}{the Conference on Empirical Methods in Natural Language Processing}.
\bibitem[{MetaAI(2024)}]{meta2024llama3}
\bibinfo{author}{MetaAI}, \bibinfo{year}{2024}.
\newblock \bibinfo{title}{Llama 3}.
\newblock \URLprefix \url{https://llama.meta.com/llama3/}. \bibinfo{note}{accessed: 2024-08-26}.
\bibitem[{Pan et~al.(2023)Pan, Shern, Zou, Li, Basart, Woodside, Ng, Zhang, Emmons and Hendrycks}]{Pan2023DoTR}
\bibinfo{author}{Pan, A.}, \bibinfo{author}{Shern, C.J.}, \bibinfo{author}{Zou, A.}, \bibinfo{author}{Li, N.}, \bibinfo{author}{Basart, S.}, \bibinfo{author}{Woodside, T.}, \bibinfo{author}{Ng, J.}, \bibinfo{author}{Zhang, H.}, \bibinfo{author}{Emmons, S.}, \bibinfo{author}{Hendrycks, D.}, \bibinfo{year}{2023}.
\newblock \bibinfo{title}{Do the rewards justify the means? measuring trade-offs between rewards and ethical behavior in the machiavelli benchmark}, in: \bibinfo{booktitle}{the International Conference on Machine Learning}.
\bibitem[{Papineni et~al.(2002)Papineni, Roukos, Ward and Zhu}]{BLEUAMethod}
\bibinfo{author}{Papineni, K.}, \bibinfo{author}{Roukos, S.}, \bibinfo{author}{Ward, T.}, \bibinfo{author}{Zhu, W.J.}, \bibinfo{year}{2002}.
\newblock \bibinfo{title}{Bleu: A method for automatic evaluation of machine translation}, in: \bibinfo{booktitle}{the Annual Meeting of the Association for Computational Linguistics}.
\newblock \URLprefix \url{https://doi.org/10.3115/1073083.1073135}.
\bibitem[{Peng et~al.(2023)Peng, Ding, Zhong, Shen, Liu, Zhang, Ouyang and Tao}]{Peng2023ChatGPT4MT}
\bibinfo{author}{Peng, K.}, \bibinfo{author}{Ding, L.}, \bibinfo{author}{Zhong, Q.}, \bibinfo{author}{Shen, L.}, \bibinfo{author}{Liu, X.}, \bibinfo{author}{Zhang, M.}, \bibinfo{author}{Ouyang, Y.}, \bibinfo{author}{Tao, D.}, \bibinfo{year}{2023}.
\newblock \bibinfo{title}{Towards making the most of chatgpt for machine translation}.
\newblock \bibinfo{journal}{arXiv preprint arXiv:2303.13780} .
\bibitem[{Pu et~al.(2023)Pu, Gao and Wan}]{Pu2023SummarizationI}
\bibinfo{author}{Pu, X.}, \bibinfo{author}{Gao, M.}, \bibinfo{author}{Wan, X.}, \bibinfo{year}{2023}.
\newblock \bibinfo{title}{Summarization is (almost) dead}.
\newblock \bibinfo{journal}{arXiv preprint arXiv:2309.09558} \bibinfo{volume}{abs/2309.09558}.
\bibitem[{Research(2019)}]{faiss2019}
\bibinfo{author}{Research, F.A.}, \bibinfo{year}{2019}.
\newblock \bibinfo{title}{Faiss}.
\newblock \URLprefix \url{https://ai.meta.com/tools/faiss/}. \bibinfo{note}{accessed: 2024-08-26}.
\bibitem[{Robertson et~al.(2009)Robertson, Zaragoza et~al.}]{robertson2009probabilistic}
\bibinfo{author}{Robertson, S.}, \bibinfo{author}{Zaragoza, H.}, et~al., \bibinfo{year}{2009}.
\newblock \bibinfo{title}{the probabilistic relevance framework: Bm25 and beyond}.
\newblock \bibinfo{journal}{Foundations and Trends{\textregistered} in Information Retrieval} .
\bibitem[{Rubin and Berant(2024)}]{rubin2023long}
\bibinfo{author}{Rubin, O.}, \bibinfo{author}{Berant, J.}, \bibinfo{year}{2024}.
\newblock \bibinfo{title}{Long-range language modeling with self-retrieval}, in: \bibinfo{booktitle}{Transactions of the Association for Computational Linguistics}.
\bibitem[{Shuster et~al.(2022)Shuster, Xu, Komeili, Ju, Smith, Roller, Ung, Chen, Arora, Lane, Behrooz, Ngan, Poff, Goyal, Szlam, Boureau, Kambadur and Weston}]{Shuster2022BlenderBot3A}
\bibinfo{author}{Shuster, K.}, \bibinfo{author}{Xu, J.}, \bibinfo{author}{Komeili, M.}, \bibinfo{author}{Ju, D.}, \bibinfo{author}{Smith, E.M.}, \bibinfo{author}{Roller, S.}, \bibinfo{author}{Ung, M.}, \bibinfo{author}{Chen, M.}, \bibinfo{author}{Arora, K.}, \bibinfo{author}{Lane, J.}, \bibinfo{author}{Behrooz, M.}, \bibinfo{author}{Ngan, W.}, \bibinfo{author}{Poff, S.}, \bibinfo{author}{Goyal, N.}, \bibinfo{author}{Szlam, A.}, \bibinfo{author}{Boureau, Y.L.}, \bibinfo{author}{Kambadur, M.}, \bibinfo{author}{Weston, J.}, \bibinfo{year}{2022}.
\newblock \bibinfo{title}{Blenderbot 3: a deployed conversational agent that continually learns to responsibly engage}.
\newblock \bibinfo{journal}{arXiv preprint arXiv:2208.03188} .
\bibitem[{Touvron et~al.(2023)Touvron, Martin, Stone, Albert, Almahairi, Babaei, Bashlykov, Batra, Bhargava, Bhosale et~al.}]{llama2}
\bibinfo{author}{Touvron, H.}, \bibinfo{author}{Martin, L.}, \bibinfo{author}{Stone, K.}, \bibinfo{author}{Albert, P.}, \bibinfo{author}{Almahairi, A.}, \bibinfo{author}{Babaei, Y.}, \bibinfo{author}{Bashlykov, N.}, \bibinfo{author}{Batra, S.}, \bibinfo{author}{Bhargava, P.}, \bibinfo{author}{Bhosale, S.}, et~al., \bibinfo{year}{2023}.
\newblock \bibinfo{title}{Llama 2: Open foundation and fine-tuned chat models}.
\newblock \bibinfo{journal}{arXiv preprint arXiv:2307.09288} .
\bibitem[{Wu et~al.(2023)Wu, Wang, Wan, Jiao and Lyu}]{wu2023chatgpt}
\bibinfo{author}{Wu, H.}, \bibinfo{author}{Wang, W.}, \bibinfo{author}{Wan, Y.}, \bibinfo{author}{Jiao, W.}, \bibinfo{author}{Lyu, M.}, \bibinfo{year}{2023}.
\newblock \bibinfo{title}{Chatgpt or grammarly? evaluating chatgpt on grammatical error correction benchmark}.
\newblock \bibinfo{journal}{arXiv preprint arXiv:2303.13648} .
\bibitem[{Wu et~al.(2022)Wu, Lan, Qian, Gu, Geramifard and Yu}]{wu-etal-2022-memformer}
\bibinfo{author}{Wu, Q.}, \bibinfo{author}{Lan, Z.}, \bibinfo{author}{Qian, K.}, \bibinfo{author}{Gu, J.}, \bibinfo{author}{Geramifard, A.}, \bibinfo{author}{Yu, Z.}, \bibinfo{year}{2022}.
\newblock \bibinfo{title}{Memformer: A memory-augmented transformer for sequence modeling}, in: \bibinfo{booktitle}{Findings of the Annual Meeting of the Association for Computational Linguistics}.
\bibitem[{Xu et~al.(2022a)Xu, Szlam and Weston}]{xu-etal-2022-beyond}
\bibinfo{author}{Xu, J.}, \bibinfo{author}{Szlam, A.}, \bibinfo{author}{Weston, J.}, \bibinfo{year}{2022}a.
\newblock \bibinfo{title}{Beyond goldfish memory: Long-term open-domain conversation}, in: \bibinfo{booktitle}{the Annual Meeting of the Association for Computational Linguistics}.
\bibitem[{Xu et~al.(2022b)Xu, Gou, Wu, Niu, Wu, Wang and Wang}]{longtimenosee}
\bibinfo{author}{Xu, X.}, \bibinfo{author}{Gou, Z.}, \bibinfo{author}{Wu, W.}, \bibinfo{author}{Niu, Z.Y.}, \bibinfo{author}{Wu, H.}, \bibinfo{author}{Wang, H.}, \bibinfo{author}{Wang, S.}, \bibinfo{year}{2022}b.
\newblock \bibinfo{title}{Long time no see! open-domain conversation with long-term persona memory}, in: \bibinfo{booktitle}{Findings of the Annual Meeting of the Association for Computational Linguistics}.
\bibitem[{Zeng et~al.(2022)Zeng, Liu, Du, Wang, Lai, Ding, Yang, Xu, Zheng, Xia, Tam, Ma, Xue, Zhai, Chen, Zhang, Dong and Tang}]{Zeng2022GLM130BAO}
\bibinfo{author}{Zeng, A.}, \bibinfo{author}{Liu, X.}, \bibinfo{author}{Du, Z.}, \bibinfo{author}{Wang, Z.}, \bibinfo{author}{Lai, H.}, \bibinfo{author}{Ding, M.}, \bibinfo{author}{Yang, Z.}, \bibinfo{author}{Xu, Y.}, \bibinfo{author}{Zheng, W.}, \bibinfo{author}{Xia, X.}, \bibinfo{author}{Tam, W.L.}, \bibinfo{author}{Ma, Z.}, \bibinfo{author}{Xue, Y.}, \bibinfo{author}{Zhai, J.}, \bibinfo{author}{Chen, W.}, \bibinfo{author}{Zhang, P.}, \bibinfo{author}{Dong, Y.}, \bibinfo{author}{Tang, J.}, \bibinfo{year}{2022}.
\newblock \bibinfo{title}{Glm-130b: An open bilingual pre-trained model}.
\newblock \bibinfo{journal}{arXiv preprint arXiv:2210.02414} .
\bibitem[{Zhang et~al.(2023a)Zhang, D’Haro, Chen, Zhang and Li}]{Zhang2023ACA}
\bibinfo{author}{Zhang, C.}, \bibinfo{author}{D’Haro, L.F.}, \bibinfo{author}{Chen, Y.}, \bibinfo{author}{Zhang, M.}, \bibinfo{author}{Li, H.}, \bibinfo{year}{2023}a.
\newblock \bibinfo{title}{A comprehensive analysis of the effectiveness of large language models as automatic dialogue evaluators}, in: \bibinfo{booktitle}{AAAI Conference on Artificial Intelligence}.
\bibitem[{Zhang et~al.(2023b)Zhang, Xiao, Liu, Dou and Nie}]{Zhang2023RetrieveAT}
\bibinfo{author}{Zhang, P.}, \bibinfo{author}{Xiao, S.}, \bibinfo{author}{Liu, Z.}, \bibinfo{author}{Dou, Z.}, \bibinfo{author}{Nie, J.Y.}, \bibinfo{year}{2023}b.
\newblock \bibinfo{title}{Retrieve anything to augment large language models}.
\newblock \bibinfo{journal}{arXiv preprint arXiv:2310.07554} .
\bibitem[{Zhang et~al.(2018)Zhang, Dinan, Urbanek, Szlam, Kiela and Weston}]{zhang-etal-2018-personalizing}
\bibinfo{author}{Zhang, S.}, \bibinfo{author}{Dinan, E.}, \bibinfo{author}{Urbanek, J.}, \bibinfo{author}{Szlam, A.}, \bibinfo{author}{Kiela, D.}, \bibinfo{author}{Weston, J.}, \bibinfo{year}{2018}.
\newblock \bibinfo{title}{Personalizing dialogue agents: {I} have a dog, do you have pets too?}, in: \bibinfo{booktitle}{the Annual Meeting of the Association for Computational Linguistics}.
\bibitem[{Zhang et~al.(2022)Zhang, Liu, Li, Zeng, Wang, You, Miao and Cui}]{2022TongZhangHierarchical}
\bibinfo{author}{Zhang, T.}, \bibinfo{author}{Liu, Y.}, \bibinfo{author}{Li, B.}, \bibinfo{author}{Zeng, Z.}, \bibinfo{author}{Wang, P.}, \bibinfo{author}{You, Y.}, \bibinfo{author}{Miao, C.}, \bibinfo{author}{Cui, L.}, \bibinfo{year}{2022}.
\newblock \bibinfo{title}{History-aware hierarchical transformer for multi-session open-domain dialogue system}, in: \bibinfo{booktitle}{Findings of the Conference on Empirical Methods in Natural Language Processing}.
\bibitem[{Zhong et~al.(2023)Zhong, Ding, Liu, Du and Tao}]{zhong2023chat}
\bibinfo{author}{Zhong, Q.}, \bibinfo{author}{Ding, L.}, \bibinfo{author}{Liu, J.}, \bibinfo{author}{Du, B.}, \bibinfo{author}{Tao, D.}, \bibinfo{year}{2023}.
\newblock \bibinfo{title}{Can chatgpt understand too? a comparative study on chatgpt and fine-tuned bert}.
\newblock \bibinfo{journal}{arXiv preprint arXiv:2302.10198} .
\bibitem[{Zhong et~al.(2024)Zhong, Guo, Gao, Ye and Wang}]{Zhong2023MemoryBankEL}
\bibinfo{author}{Zhong, W.}, \bibinfo{author}{Guo, L.}, \bibinfo{author}{Gao, Q.}, \bibinfo{author}{Ye, H.}, \bibinfo{author}{Wang, Y.}, \bibinfo{year}{2024}.
\newblock \bibinfo{title}{Memorybank: Enhancing large language models with long-term memory}, in: \bibinfo{booktitle}{Proceedings of the AAAI Conference on Artificial Intelligence}, pp. \bibinfo{pages}{19724--19731}.
\bibitem[{Zhou et~al.(2023)Zhou, Chen, Wang and Huang}]{zhou-etal-2023-facilitating}
\bibinfo{author}{Zhou, J.}, \bibinfo{author}{Chen, Z.}, \bibinfo{author}{Wang, B.}, \bibinfo{author}{Huang, M.}, \bibinfo{year}{2023}.
\newblock \bibinfo{title}{Facilitating multi-turn emotional support conversation with positive emotion elicitation: A reinforcement learning approach}, in: \bibinfo{booktitle}{Proceedings of the 61st Annual Meeting of the Association for Computational Linguistics (Volume 1: Long Papers)}, pp. \bibinfo{pages}{1714--1729}.

\end{thebibliography}


\end{document}